%% file: main.tex
\pdfoutput=1
\documentclass[10pt,twocolumn]{article}

\usepackage[margin=0.75in]{geometry}
\usepackage[T1]{fontenc}
\usepackage[utf8]{inputenc}
\usepackage{lmodern}
\usepackage{microtype}
\usepackage{amsmath,amssymb,mathtools}
\usepackage{booktabs}
\usepackage{graphicx}
\usepackage{xcolor}
\usepackage{enumitem}
\usepackage{url}
\usepackage{hyperref}
\usepackage[nameinlink,capitalize,noabbrev]{cleveref}
\usepackage{xspace}
\usepackage{caption}
\usepackage{tikz}
\usetikzlibrary{positioning,arrows.meta,calc,fit,backgrounds}

\hypersetup{colorlinks=true,linkcolor=blue!50!black,citecolor=blue!50!black,urlcolor=blue!50!black}
\hypersetup{pdftitle={Post-Training Semantic Lifting for 3D Gaussian Splatting: Separating Detector, Lifting and Representation Error},
  pdfauthor={Iván Verdugo Guerra, Ezequiel López Rubio, Jorge García González},
  pdfkeywords={3D Gaussian Splatting, semantic segmentation, label lifting, evidence accumulation, hysteresis thresholding, error decomposition}}

\makeatletter
\renewcommand\section{\@startsection{section}{1}{\z@}{-3.5ex \@plus -1ex \@minus -.2ex}{2.3ex \@plus.2ex}{\normalfont\Large\bfseries\raggedright}}
\renewcommand\subsection{\@startsection{subsection}{2}{\z@}{-3.25ex\@plus -1ex \@minus -.2ex}{1.5ex \@plus .2ex}{\normalfont\large\bfseries\raggedright}}
\makeatother

\newcommand{\gs}{3D Gaussian Splatting\xspace}
\newcommand{\mIoU}{\mathrm{mIoU}}
\newcommand{\figslot}[3]{%
  \IfFileExists{#3.pdf}%
    {\includegraphics[width=#1]{#3}}%
    {\rule{0pt}{#2}\rule{#1}{0pt}}%
}

\newcommand{\tableslot}[1]{\IfFileExists{#1.tex}{\input{#1}}{}}

\newcommand{\betaStar}{0.7}
\newcommand{\gammaStar}{0.8}
\newcommand{\tauStar}{0.10}
\newcommand{\thetaStar}{0.3}
\newcommand{\replicaGTmIoU}{XX.XX}
\newcommand{\replicaYOLOmIoU}{XX.XX}
\newcommand{\scannetGTmIoU}{XX.XX}
\newcommand{\scannetYOLOmIoU}{XX.XX}

\newcommand{\bestMean}{XX.XX}

\newcommand{\selectedMean}{XX.XX}
\newcommand{\selectedSd}{X.XX}
\newcommand{\eligibleCount}{XX}

\newcommand{\replicaDetectorGap}{XX.XX}
\newcommand{\replicaLiftingGap}{XX.XX}
\newcommand{\replicaRepGap}{XX.XX}
\newcommand{\scannetDetectorGap}{XX.XX}
\newcommand{\scannetLiftingGap}{XX.XX}
\newcommand{\scannetRepGap}{XX.XX}

\newcommand{\ablNoMtoGRef}{XX.XX}

\newcommand{\timeMasks}{XX~s}
\newcommand{\timeVotes}{XX~s}
\newcommand{\timeThreshold}{XX~s}
\newcommand{\timeTransfer}{XX~s}
\newcommand{\timeSweepWarm}{XX~s}
\newcommand{\timeMissTotal}{XX~s}

\newcommand{\memMasks}{X.X~GB}
\newcommand{\memVotes}{X.X~GB}

\newcommand{\hardwareDescription}{XXXX}

\newcommand{\scannetGTrel}{X.XXX}

\IfFileExists{macros_measured.tex}{

\input{macros_measured}}{}

\def\CapFirstAux#1#2\CapFirstEnd{\MakeUppercase{#1}#2}
\newcommand{\TransferStar}{\expandafter\CapFirstAux\transferStar\CapFirstEnd}

\newcommand{\authornote}{This preprint contains the research carried out for the Bachelor's Thesis of Iván Verdugo Guerra at the University of Málaga, which was supervised by Ezequiel López Rubio and Jorge García González. Note that the contribution of each author can be found in the Author Contributions section, on page~\pageref{sec:contributions}.}

\title{Post-Training Semantic Lifting for 3D Gaussian Splatting: Separating Detector, Lifting and Representation Error}

\author{%
\begin{tabular}{c}
Iván Verdugo Guerra\thanks{\authornote} \quad Ezequiel López Rubio \quad Jorge García González\\
\small Department of Computer Languages and Computer Science\\
\small University of Málaga\\
\texttt{\small ivanver@uma.es \quad elr@uma.es \quad jorgegarcia@uma.es}
\end{tabular}%
}

\date{October 2026\\[2pt]\small Preprint\\[2pt]
\small Code: \href{https://github.com/ivanver02/semantic-lifting-3dgs}{GitHub repository}}

\begin{document}
\twocolumn[{%
\renewcommand\twocolumn[1][]{#1}%
\maketitle
\vspace{-1.2em}
\begin{center}
\captionsetup{type=figure}
\figslot{\textwidth}{0.25\textwidth}{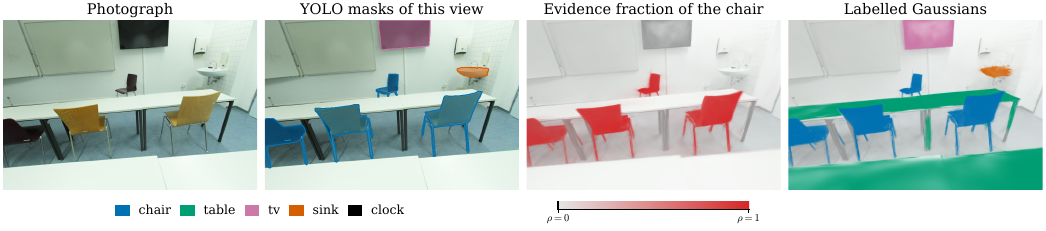}
\caption{One view of the ScanNet++ test scene \overviewScene{}. Even though YOLO does not detect
the tables in this view, they are still labelled, since it detects them in other views.}

\label{fig:overview}
\end{center}
\vspace{0.8em}
}]
{\renewcommand{\thefootnote}{\fnsymbol{footnote}}\footnotetext[1]{\authornote}}

\begin{abstract}
The same Gaussian of a 3D Gaussian Splatting model is seen from many views, and these views do not always
agree on the class it belongs to. The Gaussian may be occluded in some of them, and the confidence of the
detector is not the same from one view to another.
The ground truth, on the other hand, is given as an annotated mesh, because two training runs do not produce the same Gaussians.
In this work, we propose a post-training lifting method that works with one target class at a time and
combines the information coming from all the views. Target and non-target evidence are accumulated simultaneously,
weighted by the visibility of each Gaussian in each view. After that, the Gaussians are filtered with
two thresholds: a main threshold $\beta$ selects the high-confidence seeds,
and a lower one $\gamma\beta$ adds the connected components around them.
For the evaluation, the labels are transferred from the Gaussians to the mesh vertices that are both visible and annotated.
With this design, we can separate three sources of error: the 2D detector, the lifting and the transfer
between representations. The thresholds and the transfer operator are chosen on seven Replica
validation scenes, and the method is evaluated on ten held-out ScanNet++ scenes with the same values
for every scene and class. The mean mIoU on the validation scenes was \replicaGTmIoU{} with masks from the
dataset annotations and \replicaYOLOmIoU{} with YOLO masks, and on the ScanNet++ test scenes it was
\scannetGTmIoU{} and \scannetYOLOmIoU{}.
Compared with thresholding the evidence per view, as a previous version of the method did, the
fraction improves the test mIoU by \baseGainScannetGT{} and makes it possible to use a single threshold for all the classes and scenes of both datasets. Finally, the error analysis shows that most of the remaining error comes from the detector.
\end{abstract}


\input{sections/introduction}
\input{sections/related_work}
\input{sections/method}
\input{sections/implementation}
\input{sections/evaluation_protocol}
\input{sections/experimental_design}

\input{sections/results}
\input{sections/limitations}
\input{sections/conclusion}
\input{sections/contributions}

{\small
\bibliographystyle{plain}
\bibliography{references}
}

\appendix
\input{sections/appendix}
\end{document}

%% file: macros_measured.tex
\providecommand{\ablNoGtoMDelta}{}\renewcommand{\ablNoGtoMDelta}{-0.15}

\providecommand{\ablNoHystDelta}{}\renewcommand{\ablNoHystDelta}{+0.01}

\providecommand{\ablNoMtoGRef}{}\renewcommand{\ablNoMtoGRef}{0.76}

\providecommand{\baseCIHighScannetGT}{}\renewcommand{\baseCIHighScannetGT}{+0.28}
\providecommand{\baseCIHighScannetYOLO}{}\renewcommand{\baseCIHighScannetYOLO}{+0.17}

\providecommand{\baseCILowScannetGT}{}\renewcommand{\baseCILowScannetGT}{+0.20}
\providecommand{\baseCILowScannetYOLO}{}\renewcommand{\baseCILowScannetYOLO}{+0.06}

\providecommand{\baseClassBetaMax}{}\renewcommand{\baseClassBetaMax}{0.2}
\providecommand{\baseClassBetaMin}{}\renewcommand{\baseClassBetaMin}{0.07}

\providecommand{\baseDeltaScannetGT}{}\renewcommand{\baseDeltaScannetGT}{+0.24}
\providecommand{\baseDeltaScannetYOLO}{}\renewcommand{\baseDeltaScannetYOLO}{+0.12}

\providecommand{\baseGainScannetGT}{}\renewcommand{\baseGainScannetGT}{0.24}
\providecommand{\baseGainScannetYOLO}{}\renewcommand{\baseGainScannetYOLO}{0.12}

\providecommand{\baseRangeHigh}{}\renewcommand{\baseRangeHigh}{0.64}
\providecommand{\baseRangeLow}{}\renewcommand{\baseRangeLow}{0.34}

\providecommand{\baseScannetGTPrec}{}\renewcommand{\baseScannetGTPrec}{0.64}
\providecommand{\baseScannetGTRec}{}\renewcommand{\baseScannetGTRec}{0.83}

\providecommand{\baseScenesScannetGT}{}\renewcommand{\baseScenesScannetGT}{10}

\providecommand{\bestMean}{}\renewcommand{\bestMean}{0.94}

\providecommand{\betaStar}{}\renewcommand{\betaStar}{0.7}
\providecommand{\candidateCount}{}\renewcommand{\candidateCount}{130}

\providecommand{\clsScannetClockRef}{}\renewcommand{\clsScannetClockRef}{0.81}

\providecommand{\clsScannetTableGT}{}\renewcommand{\clsScannetTableGT}{0.72}

\providecommand{\clsScannetTableScenes}{}\renewcommand{\clsScannetTableScenes}{9}
\providecommand{\clsScannetTableYOLO}{}\renewcommand{\clsScannetTableYOLO}{0.11}

\providecommand{\clsScannetTableZero}{}\renewcommand{\clsScannetTableZero}{5}

\providecommand{\costScenes}{}\renewcommand{\costScenes}{8}

\providecommand{\eligibleCount}{}\renewcommand{\eligibleCount}{18}
\providecommand{\gammaStar}{}\renewcommand{\gammaStar}{0.8}
\providecommand{\hardwareDescription}{}\renewcommand{\hardwareDescription}{NVIDIA A100-SXM4-40GB}
\providecommand{\hystRangeHigh}{}\renewcommand{\hystRangeHigh}{0.93}
\providecommand{\hystRangeLow}{}\renewcommand{\hystRangeLow}{0.90}

\providecommand{\maskScannetTableTwoDRec}{}\renewcommand{\maskScannetTableTwoDRec}{0.09}

\providecommand{\memMasks}{}\renewcommand{\memMasks}{0.5~GB}

\providecommand{\memTraining}{}\renewcommand{\memTraining}{1.9~GB}

\providecommand{\memVotes}{}\renewcommand{\memVotes}{0.8~GB}

\providecommand{\noHystRangeHigh}{}\renewcommand{\noHystRangeHigh}{0.93}
\providecommand{\noHystRangeLow}{}\renewcommand{\noHystRangeLow}{0.38}

\providecommand{\opCIHighScannetYOLO}{}\renewcommand{\opCIHighScannetYOLO}{+0.00}

\providecommand{\opCILowScannetYOLO}{}\renewcommand{\opCILowScannetYOLO}{-0.03}

\providecommand{\opDeltaScannetYOLO}{}\renewcommand{\opDeltaScannetYOLO}{-0.01}
\providecommand{\opGainReplicaGT}{}\renewcommand{\opGainReplicaGT}{0.03}

\providecommand{\opGainScannetGT}{}\renewcommand{\opGainScannetGT}{0.01}

\providecommand{\overviewScene}{}\renewcommand{\overviewScene}{21d970d8de}

\providecommand{\radiusBetaStar}{}\renewcommand{\radiusBetaStar}{0.9}

\providecommand{\radiusClsReplicaClockGTPrec}{}\renewcommand{\radiusClsReplicaClockGTPrec}{0.64}
\providecommand{\radiusClsReplicaClockGTRec}{}\renewcommand{\radiusClsReplicaClockGTRec}{1.00}

\providecommand{\radiusGammaStar}{}\renewcommand{\radiusGammaStar}{0.7}

\providecommand{\replicaDetectorGap}{}\renewcommand{\replicaDetectorGap}{0.28}

\providecommand{\replicaGTmIoU}{}\renewcommand{\replicaGTmIoU}{0.93}

\providecommand{\replicaLiftingGap}{}\renewcommand{\replicaLiftingGap}{0.04}
\providecommand{\replicaMaskPairs}{}\renewcommand{\replicaMaskPairs}{23}

\providecommand{\replicaMaskWins}{}\renewcommand{\replicaMaskWins}{20}

\providecommand{\replicaRepGap}{}\renewcommand{\replicaRepGap}{0.03}

\providecommand{\replicaYOLOmIoU}{}\renewcommand{\replicaYOLOmIoU}{0.65}

\providecommand{\scannetDetectorGap}{}\renewcommand{\scannetDetectorGap}{0.26}
\providecommand{\scannetGTCIHigh}{}\renewcommand{\scannetGTCIHigh}{0.83}
\providecommand{\scannetGTCILow}{}\renewcommand{\scannetGTCILow}{0.77}

\providecommand{\scannetGTmIoU}{}\renewcommand{\scannetGTmIoU}{0.80}

\providecommand{\scannetGTrel}{}\renewcommand{\scannetGTrel}{0.876}
\providecommand{\scannetLiftingGap}{}\renewcommand{\scannetLiftingGap}{0.11}
\providecommand{\scannetMaskPairs}{}\renewcommand{\scannetMaskPairs}{39}

\providecommand{\scannetMaskThreeD}{}\renewcommand{\scannetMaskThreeD}{0.55}
\providecommand{\scannetMaskTwoD}{}\renewcommand{\scannetMaskTwoD}{0.45}
\providecommand{\scannetMaskTwoDPrec}{}\renewcommand{\scannetMaskTwoDPrec}{0.70}
\providecommand{\scannetMaskTwoDRec}{}\renewcommand{\scannetMaskTwoDRec}{0.51}
\providecommand{\scannetMaskWins}{}\renewcommand{\scannetMaskWins}{27}

\providecommand{\scannetRepGap}{}\renewcommand{\scannetRepGap}{0.09}

\providecommand{\scannetYOLOmIoU}{}\renewcommand{\scannetYOLOmIoU}{0.54}

\providecommand{\selectedMean}{}\renewcommand{\selectedMean}{0.93}
\providecommand{\selectedSd}{}\renewcommand{\selectedSd}{0.01}
\providecommand{\tauStar}{}\renewcommand{\tauStar}{0.10}
\providecommand{\thetaStar}{}\renewcommand{\thetaStar}{0.3}
\providecommand{\timeMasks}{}\renewcommand{\timeMasks}{103~s}
\providecommand{\timeMissTotal}{}\renewcommand{\timeMissTotal}{2463~s}
\providecommand{\timePrepare}{}\renewcommand{\timePrepare}{19.5~s}
\providecommand{\timeSweepWarm}{}\renewcommand{\timeSweepWarm}{272~s}
\providecommand{\timeThreshold}{}\renewcommand{\timeThreshold}{6.4~s}
\providecommand{\timeTraining}{}\renewcommand{\timeTraining}{617~s}
\providecommand{\timeTransfer}{}\renewcommand{\timeTransfer}{257~s}
\providecommand{\timeVotes}{}\renewcommand{\timeVotes}{1359~s}
\providecommand{\transferStar}{}\renewcommand{\transferStar}{nearest Gaussian}

%% file: sections/introduction.tex
\section{Introduction}
\label{sec:introduction}

\gs represents a reconstructed scene as a set of anisotropic 3D Gaussians
\cite{kerbl2023gaussian,kerbl3dgsrepo}. Each of them stores its own geometry and appearance,
that is, its position, covariance, opacity and spherical harmonics.
The class to which the Gaussian belongs, however, is not part of these parameters.

In most cases, the semantic information comes from the 2D masks that a detector produces on the images used for the reconstruction.
These masks are projected back onto the Gaussians, and the main problem is that the same Gaussian is seen by several
cameras that do not always agree: it may be occluded in one view, or detected with a different confidence in another.
There is also a second difficulty. The labels are evaluated on a mesh, which is a set of fixed vertices, while the Gaussian
centres form a sparse cloud. As a consequence, a labelling that looks correct on the Gaussians can still
lose accuracy when it is transferred to the mesh, since the area that a primitive covers in a view depends on its
anisotropic shape and on its depth.

In general, the methods that add semantics to a Gaussian scene can be divided into two groups. Some of them train semantic
features into the primitives themselves, either during or after the reconstruction
\cite{lerf2023,langsplat2024,gaussiangrouping2024,opengaussian2024}, and others lift the labels
onto the Gaussians once the model is trained \cite{flashsplat2024,ludvig2025,cdseg2026}.
The method that we present in this paper belongs to the second group.

In our case, the masks are the only semantic input of the method. For each camera, the means and
covariances of the visible Gaussians are projected \cite{zwicker2002ewa}
and their contribution is weighted with alpha compositing in depth order. In the same pass, two quantities are accumulated for every Gaussian: the evidence that it belongs to the target class and the evidence that it belongs
to anything else. The ratio between the first one and the sum of both is what we call the target evidence fraction, and it is
thresholded at a fixed $\beta$. A Gaussian below that threshold but above $\gamma\beta$ still gets the label of the
target class when it belongs to the same connected component as a seed. Finally, for the evaluation, the labelled
Gaussians are transferred to the mesh vertices, where they are compared with the ground truth.
Figure~\ref{fig:overview} shows an example of the method on one view of a test scene.

The main aim of our evaluation is to find out in which part of the pipeline the accuracy is lost. YOLO masks \cite{yolo26} are not the same as the masks derived from the dataset annotations, and this is something that we can take advantage of: running the same method with both sources allows us to separate the detector from the rest of the pipeline. With the
annotation masks, whatever is still lost comes either from the lifting or from
the representation, because a mesh vertex may not be reachable from any Gaussian, or may be
reached mostly by Gaussians of the wrong class. This second part is measured with
a reference that sends the annotations through the same transfer operator.

The development of the method was carried out on \texttt{office\_0} from Replica \cite{replica19arxiv} and \texttt{7831862f02} from
ScanNet++ \cite{yeshwanth2023scannetpp}, and both scenes were left out of validation
and test. The pair $(\beta, \gamma)$ and the operator that carries the labels to the mesh are selected with a
sweep over seven Replica validation scenes, and ten held-out ScanNet++ scenes are then evaluated with the same parameters.
Both mask sources are used in both datasets.

The main contributions of this work are the following:
\begin{itemize}[leftmargin=*,itemsep=2pt]
  \item A semantic voting scheme in which the visibility is taken into account. Target
        and non-target evidence are accumulated at the same time, and a spatial hysteresis makes the result more stable.
  \item A PyTorch implementation by tiles of the visibility computation of the official
        Gaussian Splatting renderer \cite{kerbl3dgsrepo}, adapted to semantic votes weighted by
        confidence. All the differences with respect to the CUDA code are documented.
  \item A selection protocol that fixes the pair $(\beta,\gamma)$ and the transfer operator from the
        validation mean and the variation between scenes, with a tolerance of $0.01$ mIoU.
  \item An error analysis that separates 2D detection, lifting and representation, and two analyses that
        explain the result: the evidence fraction against the evidence per view of a previous version,
        which it improves by \baseGainScannetGT{} mIoU on the test scenes, and the 3D result against the
        2D quality of the detector. All of this comes with a container workflow for
        Docker and Apptainer and cached intermediate data to reproduce the experiments.
\end{itemize}

The rest of the paper is organised as follows. Section~\ref{sec:related} reviews the related work,
Section~\ref{sec:method} describes the method and Section~\ref{sec:implementation} its implementation.
Sections~\ref{sec:setup} and~\ref{sec:design} present the evaluation protocol and the experimental design,
and Section~\ref{sec:results} reports the results. The paper ends with the limitations of the method and
the conclusions of the work, in Sections~\ref{sec:limitations} and~\ref{sec:conclusion}.

%% file: sections/related_work.tex
\section{Related Work}
\label{sec:related}

\subsection{Gaussian scene representations}
\label{sec:related-gaussian}

Since the appearance of 3D Gaussian Splatting, several works have proposed changes in the geometry of the primitive. For example, 2D Gaussian Splatting replaces
the ellipsoid with an oriented planar disk. A ray meets a disk at a single
point, so the depth and the normal do not change with the viewpoint, and the resulting
geometry is more accurate than with volumetric primitives \cite{huang2024two}.
Our implementation is based on the official release \cite{kerbl3dgsrepo}, and this is why it keeps the original Gaussians and the same training procedure.

Other methods include the semantic information in the representation itself, during the optimisation.
Semantic-NeRF learns an implicit semantic field together with the radiance field \cite{zhi2021semantic}.
This is not only done with NeRF, as Gaussian models are also trained in this way, either with language features
for text queries \cite{lerf2023,langsplat2024}, or with identity and affinity features that
group the primitives into objects \cite{gaussiangrouping2024,saga2025,opengaussian2024}.
In all these cases, the semantic label is learnt together with the model itself.

\subsection{Label lifting from images to 3D}
\label{sec:related-lifting}

The idea of transferring image predictions to a common 3D representation is not new, and it is in fact older than Gaussian scenes.
In RGB-D pipelines, the predictions of each frame are accumulated during the reconstruction
with Bayesian updates and pairwise regularisation over the points \cite{hermans2014dense,mccormac2017semanticfusion}.
Closer to our work, Kundu et al.\ render a reconstructed mesh from virtual viewpoints and
aggregate the predictions of each view on its vertices \cite{kundu2020virtual}, and
Genova et al.\ reuse this aggregation to train a 3D segmentation
network from images only \cite{genova2021learning}. All of them have to deal with calibration,
visibility and views that contradict each other, and they differ mainly in the weight
given to each observation. In our case, this weight comes from the projected covariance of the Gaussian
and from its contribution to the pixel.

An aggregated label usually accumulates errors from several sources: the images,
the aggregation step and the representation where it ends up. To study them,
Kundu et al.\ \cite{kundu2020virtual} compare the accuracy of the predictions of each
view with the accuracy on the mesh where they are aggregated.
Semantic-NeRF, on the other hand, is trained on labels that have been corrupted on purpose,
which shows how much of that noise is absorbed by the consistency between views
\cite{zhi2021semantic}. To measure the cost of the representation alone, we send
the ground-truth annotation through the same transfer operator as the prediction. A similar idea
has already been used in LiDAR segmentation, where each point takes the label of its
cell and the resulting mIoU is interpreted as the upper bound imposed by the grid
\cite{zhang2020polarnet,zhou2021panopticpolarnet}, and also with
superpixels in indoor RGB-D segmentation \cite{he2017std2p}. As our transfer operator has
its own parameters, we report this value together with the settings that produced it.

\subsection{Post-training label lifting on Gaussian models}
\label{sec:related-posttraining}

The closest work to ours is CDSeg, submitted on 6 August 2026 \cite{cdseg2026}. It builds
a Gaussian carrier that can be rendered, either by completing each input
point with a single Gaussian and keeping its index, or by reusing the primitives of
a Gaussian scene that is already optimised. After that, it records the associations between pixels and primitives
given by the renderer, aggregates the mask labels by voting and
applies a local filter. The setting is similar to ours, as in both cases the labels go from the images to the primitives after the reconstruction, using the visibility of the renderer.
Nevertheless, the goal of both works is different. CDSeg is a general interface for both instance and semantic
tasks, which also handles LiDAR points, and its main priorities are keeping the indices and building the carrier.
We work instead with one class at a time, we weight each vote
by the confidence of the detector and we let target and non-target evidence compete
in the same pass. We also measure the cost of the transfer to an annotated mesh.

The rest of the post-training methods differ mainly in the way in which the information of the images
is assigned to each primitive. FlashSplat solves the assignment as a global linear
problem with a background bias \cite{flashsplat2024}, while LUDVIG aggregates DINO,
SAM or CLIP features in the scene and refines them with graph diffusion
\cite{ludvig2025}. Finally, PointGS densifies a point cloud into a
Gaussian space, distils the semantics there and
registers the labels back onto the input points \cite{pointgs2026}.

\subsection{Thresholding and spatial consistency}
\label{sec:related-hysteresis}

Applying a threshold to a score and then using spatial information to correct the result is a very common
idea in computer vision, and it has been used for many years. For example, the Canny edge detector keeps a
weak response when it is connected to a strong one \cite{canny1986edge}. Our spatial
hysteresis does the same with two thresholds, but on a graph of Gaussians instead of on an image.
Other approaches in this line are dense CRFs, which regularise the pixel labels with pairwise potentials
over the whole image \cite{krahenbuhl2011efficient}, and graph-based segmentation, which
merges regions according to a predicate on the edge weights \cite{felzenszwalb2004efficient}.
In our case, the approach is simpler, as we only use one connectivity radius and one low threshold.

%% file: sections/method.tex
\section{Method}
\label{sec:method}

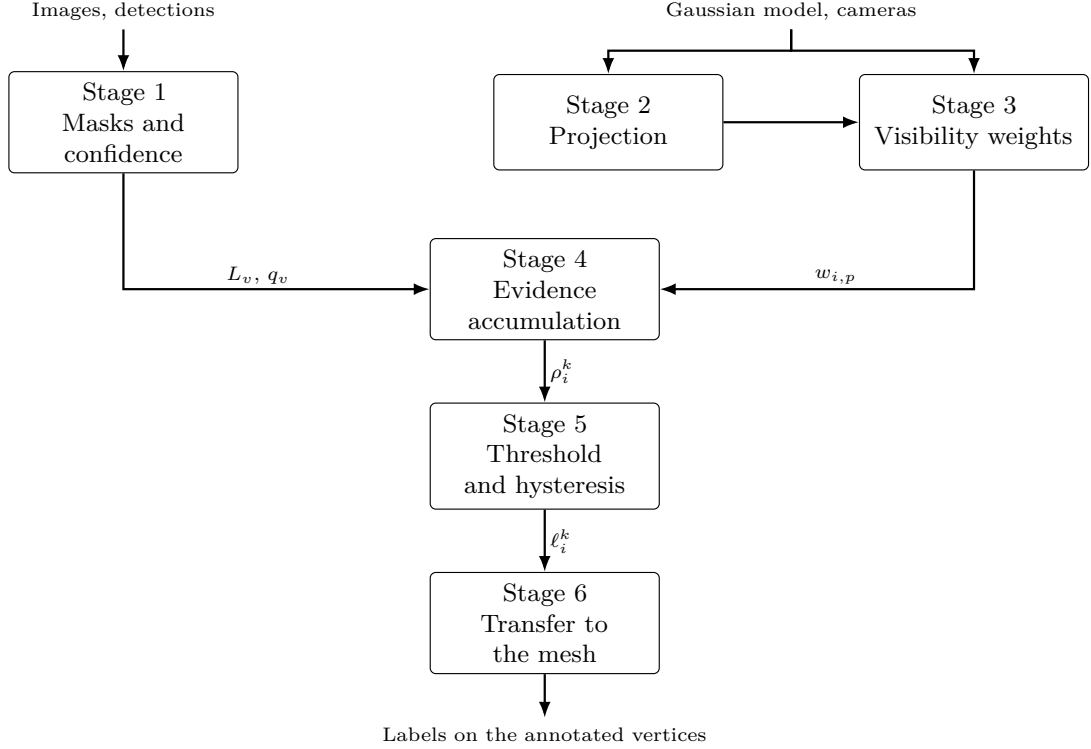
\begin{figure*}[t]
\centering
\resizebox{0.82\textwidth}{!}{%
\begin{tikzpicture}[
  box/.style={draw, rounded corners=2pt, align=center, inner sep=4pt,
              text width=26mm, font=\small, minimum height=12mm},
  io/.style={align=center, font=\scriptsize, text width=32mm},
  lab/.style={font=\scriptsize, inner sep=1.5pt},
  flow/.style={-{Latex[length=2mm]}, thick}
]
\node[box] (p1) at (1.9,3.1) {Stage 1\\Masks and confidence};
\node[box] (p2) at (8.0,3.1) {Stage 2\\Projection};
\node[box] (p3) at (12.6,3.1) {Stage 3\\Visibility weights};
\node[box] (p4) at (7.2,1.0)  {Stage 4\\Evidence accumulation};
\node[box] (p5) at (7.2,-1.1) {Stage 5\\Threshold and hysteresis};
\node[box] (p6) at (7.2,-3.2) {Stage 6\\Transfer to the mesh};

\node[io] (in1) at (1.9,4.5) {Images, detections};
\node[io] (in2) at (10.3,4.5) {Gaussian model, cameras};
\node[io, text width=48mm] (out) at (7.2,-4.6)
      {Labels on the annotated vertices};

\draw[flow] (in1) -- (p1);
\draw[flow] (in2) -- ++(0,-0.5) -| (p2);
\draw[flow] (in2) -- ++(0,-0.5) -| (p3);
\draw[flow] (p2) -- (p3);
\draw[flow] (p1.south) |- node[lab, pos=0.72, above] {$L_v$, $q_v$} (p4.west);
\draw[flow] (p3.south) |- node[lab, pos=0.72, above] {$w_{i,p}$} (p4.east);
\draw[flow] (p4) -- node[lab, right] {$\rho_i^{k}$} (p5);
\draw[flow] (p5) -- node[lab, right] {$\ell_i^{k}$} (p6);
\draw[flow] (p6) -- (out);
\end{tikzpicture}}
\caption{The six stages of the method. Stages 2 and 3 do not depend on the class, and a change of
$\beta$ or $\gamma$ only runs the pipeline again from stage 5.}
\label{fig:pipeline}
\end{figure*}

\subsection{Problem formulation}
\label{sec:method-problem}

Let $\mathcal{G}=\{g_i\}_{i=1}^{N}$ be a Gaussian model that has already been optimised for
photometric reconstruction from a set of calibrated cameras $\mathcal{V}$. Each Gaussian $g_i$ is defined by its mean
$\boldsymbol{\mu}_i\in\mathbb{R}^{3}$, a rotation $\mathbf{R}_i\in SO(3)$, a
diagonal scale $\mathbf{S}_i$ and an opacity $o_i\in[0,1]$. Its covariance, which
is the one used during the optimisation \cite{kerbl2023gaussian}, is obtained from the rotation and the scale as
\begin{equation}
  \boldsymbol{\Sigma}_i
  =\mathbf{R}_i\mathbf{S}_i\mathbf{S}_i^{\top}\mathbf{R}_i^{\top}.
  \label{eq:covariance}
\end{equation}

The segmentation is posed as a one-vs-rest problem, which is solved once for each target class of the dataset taxonomy. Given a class $k$, every camera provides a 2D mask that
separates target and non-target pixels, with a confidence for each pixel. From
these masks, the method assigns a binary label $\ell_i^{k}\in\{0,1\}$ to each
Gaussian, and these labels are then transferred to the mesh vertices of the scene for the evaluation,
as described in Section~\ref{sec:method-transfer}.

During the projection, two non-negative quantities are accumulated for each Gaussian:
the target evidence $E_i^{+}$ and the non-target evidence $E_i^{-}$,
both with the same visibility weights of Eq.~\eqref{eq:alpha}. Their sum is zero only when the Gaussian has no
contributing view, that is, when it lies behind the near plane or its projected centre falls outside the
image in every view where the class appears. Occlusion alone does not make the sum zero,
because small contributions are not discarded at this point. The decision is made on the support index set
\begin{equation}
  \mathcal{S}^{k}=\bigl\{\,i\in\{1,\dots,N\}\;:\;E_i^{+}+E_i^{-}>0\,\bigr\}.
  \label{eq:support}
\end{equation}
Gaussians outside $\mathcal{S}^{k}$ keep $\ell_i^{k}=0$. Inside it, the label of
$g_i$ depends on the pair $(E_i^{+},E_i^{-})$ and on the other Gaussians of
$\mathcal{S}^{k}$ that lie within a fixed radius of $\boldsymbol{\mu}_i$.

The six stages of the method are shown in Figure~\ref{fig:pipeline}, and the following subsections
describe them in the order in which they run.

\subsection{Two-dimensional masks and confidence}
\label{sec:method-masks}

Each image provides two inputs to the method: a label map
$L_v$, whose value at each pixel is the identifier stored for that pixel, and a detector
confidence map $q_v$ with values in $[0,1]$. If the dimensions do not match,
the detector masks are resampled to the camera resolution with nearest-neighbour interpolation.
The masks obtained from the datasets are written in the same format, with
confidence one on every annotated pixel, so the following stages treat
both sources in the same way.

For the target class $k$, the pixel grid $\Omega_v$ of view $v$ is split into
\begin{equation}
  \mathcal{P}_v^{+}=\{\,p\in\Omega_v: L_v(p)=k\,\},
  \qquad
  \mathcal{P}_v^{-}=\Omega_v\setminus\mathcal{P}_v^{+},
  \label{eq:pixelsets}
\end{equation}
where the identifier zero is reserved for pixels without any detection. These
pixels also belong to $\mathcal{P}_v^{-}$, but they do not have a confidence of
their own, so
\begin{equation}
  q_v(p)=
  \begin{cases}
    \hat{q}_v(p), & L_v(p)\neq 0,\\[2pt]
    q_{0},        & L_v(p)=0,
  \end{cases}
  \label{eq:confidence}
\end{equation}
where $\hat{q}_v(p)=s_v(p)\,c$ is the value of the soft mask $s_v$ at pixel $p$ multiplied by the
score $c$ of the whole detection. In this way, the confidence changes inside the same object mask.
The value $q_{0}=0.25$ is a fixed confidence that is given to the background pixels.
Therefore, a pixel with a detected label keeps the confidence of its detection, whether
that label is the target or not. At first, the same confidence was given to every pixel of a detection,
which was simpler, and then we let it vary per pixel with the soft mask, which slightly improved the results.

\subsection{Projection of means and covariances}
\label{sec:method-projection}

Given a Gaussian $g_i$, its mean $\boldsymbol{\mu}_i \in \mathbb{R}^{3}$ is first
transformed to camera coordinates $\boldsymbol{\mu}_i^{\text{cam}} = (x_i, y_i, z_i)^{\top}$ with the rigid transform
\begin{equation}
  \boldsymbol{\mu}_i^{\text{cam}} = \mathbf{W}\boldsymbol{\mu}_i + \mathbf{t},
\end{equation}
and only the Gaussians in front of the near plane, that is, with $z_i > 0$, are kept. For a pinhole camera,
the projected 2D centre $\boldsymbol{\mu}'_i = (u_i, v_i)^{\top}$ is
\begin{equation}
  u_i = f_x\frac{x_i}{z_i} + c_x,
  \qquad
  v_i = f_y\frac{y_i}{z_i} + c_y,
  \label{eq:projection}
\end{equation}
where $(c_x,c_y)$ is the principal point given by the camera calibration.

The covariance is projected with the local affine approximation of EWA splatting
\cite{zwicker2002ewa}, in the same form as in the Gaussian Splatting rasteriser
\cite{kerbl2023gaussian}:
\begin{equation}
  \boldsymbol{\Sigma}'_i = \mathbf{J}_i\,\mathbf{W}\,\boldsymbol{\Sigma}_i\,\mathbf{W}^{\top}\mathbf{J}_i^{\top},
  \label{eq:cov2d}
\end{equation}
where $\mathbf{W}$ is the rotation part of the world-to-camera transform and
$\mathbf{J}_i$ is the Jacobian of the projection evaluated at
$\boldsymbol{\mu}_i^{\text{cam}}$,
\begin{equation}
  \mathbf{J}_i =
  \begin{bmatrix}
    f_x/z_i & 0       & -f_x x_i/z_i^{2}\\[2pt]
    0       & f_y/z_i & -f_y y_i/z_i^{2}
  \end{bmatrix}.
  \label{eq:jacobian}
\end{equation}

As in the standard rasteriser \cite{kerbl2023gaussian}, a low-pass filter of
$0.3\,\text{px}^2$ is added to the projected covariance, that is, we compute
$\boldsymbol{\Sigma}'_i+0.3\,\mathbf{I}_{2}$ before the inversion. In this way,
the covariance of the Gaussians smaller than a pixel can still be inverted without numerical problems.
The inverse is symmetric and is stored as three conic coefficients $(A, B, C)$,
so the exponent at an offset $(\Delta u, \Delta v)$ becomes the quadratic
form $-\frac{1}{2}(A\Delta u^2 + C\Delta v^2) - B\Delta u\Delta v$ and no matrix has to be
inverted per pixel. From the largest eigenvalue $\lambda_i$ of the filtered
covariance, we obtain the conservative radius
\begin{equation}
  r_i=\bigl\lceil 3\sqrt{\lambda_i}\bigr\rceil,
  \label{eq:radius}
\end{equation}
which bounds, in whole pixels, the region that $g_i$ can reach.
The Gaussians whose projected centre falls outside the image are discarded, as Section~\ref{sec:method-raster} explains.

\subsection{Tile rasterisation and visibility weights}
\label{sec:method-raster}

As in the original rasteriser, the image is divided into square tiles of 16 pixels per side. Each remaining
Gaussian is assigned to every tile that its radius intersects, and inside a tile
all the relevant pairs of Gaussian and pixel are evaluated at once with tensor broadcasting.

For Gaussian $i$ and pixel $p$ at position $\mathbf{x}_p$, the alpha value is
the projected Gaussian multiplied by the opacity,
\begin{equation}
  \alpha_{i,p}
  =o_i\exp\!\left(-\tfrac{1}{2}
   (\mathbf{x}_p-\boldsymbol{\mu}'_i)^{\top}
   \bigl(\boldsymbol{\Sigma}'_i\bigr)^{-1}
   (\mathbf{x}_p-\boldsymbol{\mu}'_i)\right),
  \label{eq:alphavalue}
\end{equation}
and the Gaussians that cover $p$ are composited from front to back after sorting them
by depth \cite{max1995optical}:
\begin{equation}
  T_{i,p}=\prod_{j<i}(1-\alpha_{j,p}),
  \qquad
  w_{i,p}=\alpha_{i,p}T_{i,p},
  \label{eq:alpha}
\end{equation}
where $j<i$ runs over the Gaussians that cover $p$ and are closer to the camera.
When rendering, $T_{i,p}$ is the transmittance that reaches $g_i$ and
$w_{i,p}$ its contribution to the colour of $p$. The accumulation of
Section~\ref{sec:method-votes} uses $w_{i,p}$ as the weight of the semantic
evidence that pixel $p$ gives about Gaussian $i$. As a result, a Gaussian that is hidden behind closer ones receives almost no weight from that view.

This visibility computation follows, with some changes, the CUDA rasteriser released with 3D
Gaussian Splatting \cite{kerbl2023gaussian,kerbl3dgsrepo}, but it is written in Python
and PyTorch. The main difference with respect to the original implementation is in the culling step. The culling of
Section~\ref{sec:method-projection} is stricter than in the reference, which
keeps any Gaussian whose radius intersects a tile. We decided to change it because we observed that a few very large Gaussians, with their centres outside the frustum,
collected target evidence from several views through the tiles that their
support touched. For this reason, in our implementation the centre of a Gaussian
must lie inside the frustum, even if its radius intersects a tile.
The reference also cuts the compositing loop short. It discards a contribution once its
alpha is too small to be visible, and it stops processing a pixel once its accumulated
transmittance is negligible. In practice, we do not apply either of these cuts, so the nearly transparent
contributions also reach Equation~\eqref{eq:evidence}, and an opaque Gaussian
centred on a pixel takes the transmittance of that pixel to zero.

\subsection{Target and non-target evidence}
\label{sec:method-votes}

Both channels are filled in the same pass over the cameras. A camera
contributes to class $k$ when a mask was matched to it and that mask contains at least one
pixel of the class,
\begin{equation}
  \mathcal{V}^{k}=\{\,v \in \mathcal{V}: \mathcal{P}_v^{+}\neq\emptyset\,\},
  \label{eq:views}
\end{equation}
and using the pixel sets of Equation~\eqref{eq:pixelsets}, the confidence of
Equation~\eqref{eq:confidence} and the visibility weight of
Equation~\eqref{eq:alpha},
\begin{equation}
  E_i^{\pm}=\sum_{v\in\mathcal{V}^{k}}\;\sum_{p\in\mathcal{P}_v^{\pm}}
            q_v(p)\,w_{i,p}.
  \label{eq:evidence}
\end{equation}

The weight $w_{i,p}$ is computed only once per view and is used by both sums, so target
and non-target evidence see the same occlusions and the same visibility, and
the only thing that separates the two channels is the mask. A view where the class does
not appear is skipped before any accumulation, so it contributes to neither channel.
Otherwise, the cameras that never see the class would dilute the evidence, and the meaning
of the $\beta$ threshold, introduced in Section~\ref{sec:method-hysteresis}, would change
from one class to another.

The target evidence fraction of Gaussian $i$ is
\begin{equation}
  \rho_i^{k}=\frac{E_i^{+}}{E_i^{+}+E_i^{-}},
  \qquad i\in\mathcal{S}^{k},
  \label{eq:score}
\end{equation}
which lies in $[0,1]$ by construction. Note that the fraction also depends on
$\mathcal{V}^{k}$. A camera that never sees class $k$ only has non-target
pixels, and if these cameras were included, in a scene where the class is rare every
fraction would be pushed towards zero. This alternative policy, in which every camera
enters both sums, is the one reported in Section~\ref{sec:ablations}.

\subsection{Threshold and spatial hysteresis}
\label{sec:method-hysteresis}

The parameter $\beta \in (0, 1]$ is the fraction of target evidence that a Gaussian must reach by itself.
For any threshold $t \in [0, 1]$, we define the corresponding level set as
\begin{equation}
  \mathcal{S}^{k}_{t} = \bigl\{\,i \in \mathcal{S}^{k} : \rho_i^{k} \geq t\,\bigr\}.
  \label{eq:levelsets}
\end{equation}
We added a second threshold after seeing what happened with a lower $\beta$: the Gaussians of
the object came back, but so did the noise around it, and hysteresis allows us to recover the
first ones without the second. With $\gamma \in [0, 1)$, we have by construction
$\mathcal{S}^{k}_{\beta} \subseteq \mathcal{S}^{k}_{\gamma\beta}$.
The elements of $\mathcal{S}^{k}_{\beta}$ are the seeds, and the elements of
$\mathcal{S}^{k}_{\gamma\beta}$ are split into the connected components of a graph
that joins any two Gaussians within a distance $\eta$ of each other. All the Gaussians of a
component are labelled as the target class when the component contains at least one
seed. When $\gamma=0$ there is no expansion, and only the seeds get label one.

\begin{figure}[t]
\centering
\resizebox{\columnwidth}{!}{%
\definecolor{classk}{RGB}{0,114,178}%
\begin{tikzpicture}[
  font=\small,
  seed/.style={circle, draw, fill=black, inner sep=2.1pt},
  cand/.style={circle, draw, fill=white, inner sep=2.1pt},
  low/.style={circle, fill=black!25, inner sep=1.2pt},
  lab/.style={circle, draw=black!70, fill=classk, inner sep=2.1pt},
  gone/.style={circle, draw=black!45, fill=black!10, inner sep=2.1pt},
  edge/.style={black!55, thick},
  faint/.style={black!18, thick}
]
\foreach \dx in {0, 6.4}
  \foreach \p in {(3.45,1.75), (0.15,2.35), (3.65,0.15), (1.05,0.1), (4.85,2.0), (4.95,0.1)}
    \node[low, xshift=\dx cm] at \p {};
\coordinate (s1) at (0.8,1.6);  \coordinate (s2) at (2.2,1.7);
\coordinate (c1) at (0.4,0.9);  \coordinate (c2) at (1.3,2.25);
\coordinate (c3) at (1.5,1.05); \coordinate (c4) at (2.9,2.2);
\coordinate (c5) at (2.8,1.0);  \coordinate (c6) at (1.9,0.5);
\coordinate (n1) at (4.0,0.55); \coordinate (n2) at (4.6,0.95); \coordinate (n3) at (4.2,1.25);
\draw[edge] (s1)--(c1) (s1)--(c2) (s1)--(c3) (c3)--(s2) (c2)--(s2) (s2)--(c4) (s2)--(c5) (c3)--(c6) (c6)--(c5);
\draw[edge] (n1)--(n2) (n1)--(n3) (n2)--(n3);
\foreach \c in {s1,s2} \node[seed] at (\c) {};
\foreach \c in {c1,c2,c3,c4,c5,c6,n1,n2,n3} \node[cand] at (\c) {};
\node[font=\normalsize] at (2.5,3.0) {Before the hysteresis};
\draw[{Bar[width=4pt]}-{Bar[width=4pt]}] (0.2,-0.5) -- node[below=1pt] {$\eta$} ++(1.05,0);
\draw[-{Latex[length=2.4mm]}, very thick, black!60] (5.35,1.3) -- (6.2,1.3);
\begin{scope}[xshift=6.4cm]
  \draw[faint] (0.8,1.6)--(0.4,0.9) (0.8,1.6)--(1.3,2.25) (0.8,1.6)--(1.5,1.05) (1.5,1.05)--(2.2,1.7)
               (1.3,2.25)--(2.2,1.7) (2.2,1.7)--(2.9,2.2) (2.2,1.7)--(2.8,1.0) (1.5,1.05)--(1.9,0.5)
               (1.9,0.5)--(2.8,1.0);
  \foreach \p in {(0.8,1.6), (2.2,1.7), (0.4,0.9), (1.3,2.25), (1.5,1.05), (2.9,2.2), (2.8,1.0), (1.9,0.5)}
    \node[lab] at \p {};
  \foreach \p in {(4.0,0.55), (4.6,0.95), (4.2,1.25)} \node[gone] at \p {};
  \node[font=\normalsize] at (2.5,3.0) {After the hysteresis};
\end{scope}
\begin{scope}[yshift=-1.35cm]
  \node[seed] at (0.3,0) {};     \node[right] at (0.45,0) {Seed, $\rho_i^{k}\ge\beta$};
  \node[cand] at (4.3,0) {};     \node[right] at (4.45,0) {Candidate, $\gamma\beta\le\rho_i^{k}<\beta$};
  \node[low] at (0.3,-0.6) {};   \node[right] at (0.45,-0.6) {$\rho_i^{k}<\gamma\beta$};
  \node[lab] at (4.3,-0.6) {};   \node[right] at (4.45,-0.6) {Labelled as class $k$};
  \node[gone] at (8.4,-0.6) {};  \node[right] at (8.55,-0.6) {Discarded};
\end{scope}
\end{tikzpicture}
}
\caption{Hysteresis on the graph of Gaussian centres. A component takes the class only when it contains a seed.}
\label{fig:hysteresis}
\end{figure}
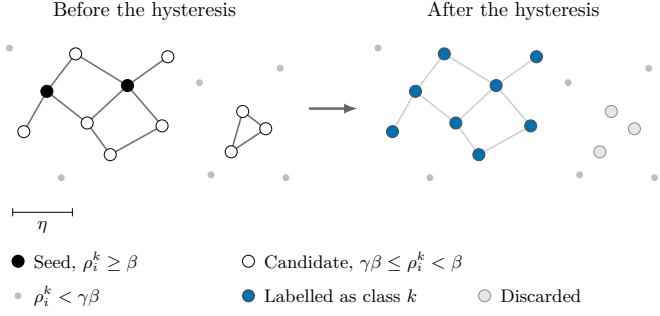

An example of this process is shown in Figure~\ref{fig:hysteresis}. The idea is the same as in the
hysteresis of edge detection \cite{canny1986edge}, but with the neighbourhood defined in the scene itself. As in Canny's method, the evidence is not propagated
between Gaussians, so Equation~\eqref{eq:score} does not change. It should be noted that the method gives one
binary label per class and does not distinguish between instances.

The radius $\eta$ was fixed on the development scenes, and the rule of
Section~\ref{sec:selection} gave $(\beta,\gamma)=(\betaStar,\gammaStar)$ on
the seven Replica validation scenes.
In the evaluation, the same pair $(\beta,\gamma)$ is used for every class and every scene.

\subsection{Transfer from Gaussians to mesh}
\label{sec:method-transfer}

Since the prediction is evaluated on the mesh, it is necessary to transfer the selected Gaussians to it. The two operators that we consider look at the same neighbourhood. An annotated vertex
$\mathbf{v}$ collects the Gaussians whose centres are within a radius $\tau$ of it,
\begin{equation}
  \mathcal{N}(\mathbf{v}) = \bigl\{\,i \in \{1, \dots, N\} : \|\boldsymbol{\mu}_i - \mathbf{v}\| \le \tau\,\bigr\},
  \label{eq:mesh-transfer}
\end{equation}
and a vertex with an empty neighbourhood stays unlabelled.

The choice of this operator has an important effect on the results, as we will see in
Section~\ref{sec:results-operator}, so we consider two of them. The first operator is the nearest
Gaussian. The vertex takes the label of the Gaussian of $\mathcal{N}(\mathbf{v})$ whose centre is
closest to it, with no weighting. In this way, the labelled surface follows the labelled Gaussians and
nothing more.

\begin{figure}[t]
\centering
\resizebox{\columnwidth}{!}{%
\definecolor{classk}{RGB}{0,114,178}\definecolor{closest}{RGB}{213,94,0}%
\begin{tikzpicture}[
  font=\small,
  cls/.style={draw=black!70, fill=classk},
  bkg/.style={draw=black!70, fill=white},
  link/.style={black!45},
  vtx/.style={rectangle, draw, inner sep=2.4pt}
]
\foreach \dx/\title in {0/Nearest Gaussian, 6.2/Radius vote} {
  \begin{scope}[xshift=\dx cm]
    \node[font=\normalsize] at (2.5,2.4) {\title};
    \draw[thick] (0,-0.05) .. controls (1.5,0.35) and (3.4,-0.3) .. (5,0.1);
    \draw[dashed, black!60] (2.5,0.05) circle (1.5);
    \node[black!60] at (0.5,1.6) {Object};
    \node[black!60] at (4.4,1.6) {Background};
    \foreach \p/\a in {(2.15,0.35)/20, (1.8,0.75)/-30, (1.3,0.4)/10, (0.2,0.9)/35}
      \draw[cls] \p ellipse[x radius=0.24, y radius=0.14, rotate=\a];
    \foreach \p/\a in {(2.8,0.25)/-15, (3.25,0.75)/40, (2.95,-0.5)/0, (4.6,0.6)/25}
      \draw[bkg] \p ellipse[x radius=0.24, y radius=0.14, rotate=\a];
  \end{scope}
}
\draw[line width=1.8pt, closest] (2.5,0.05) -- (2.8,0.25);
\node[vtx, fill=white] at (2.5,0.05) {};
\node[below=2pt] at (2.5,-0.05) {$\mathbf{v}$};
\begin{scope}[xshift=6.2cm]
  \foreach \p in {(2.15,0.35), (1.8,0.75), (1.3,0.4), (2.8,0.25), (3.25,0.75), (2.95,-0.5)}
    \draw[link] (2.5,0.05) -- \p;
  \draw[-{Latex[length=1.6mm]}] (2.5,0.05) -- node[below right=-1pt] {$\tau$} ++(-30:1.5);
  \node[vtx, fill=classk] at (2.5,0.05) {};
  \node[below=2pt] at (2.5,-0.05) {$\mathbf{v}$};
\end{scope}
\begin{scope}[yshift=-1.9cm]
  \draw[cls] (0.3,0) ellipse[x radius=0.2, y radius=0.12];  \node[right] at (0.55,0) {Gaussian with $\ell_i^{k}=1$};
  \draw[bkg] (5.5,0) ellipse[x radius=0.2, y radius=0.12];  \node[right] at (5.75,0) {Gaussian with $\ell_i^{k}=0$};
  \node[vtx] at (0.3,-0.6) {};                              \node[right] at (0.55,-0.6) {Annotated vertex};
  \draw[line width=1.8pt, closest] (5.3,-0.6) -- (5.7,-0.6); \node[right] at (5.75,-0.6) {Closest centre};
\end{scope}
\end{tikzpicture}
}
\caption{The two transfer operators at a vertex next to the border of an object. With $\theta$ below $0.5$, the radius vote gives the class to vertices where the nearest Gaussian does not, so the labelled region grows.}
\label{fig:vertexvote}
\end{figure}
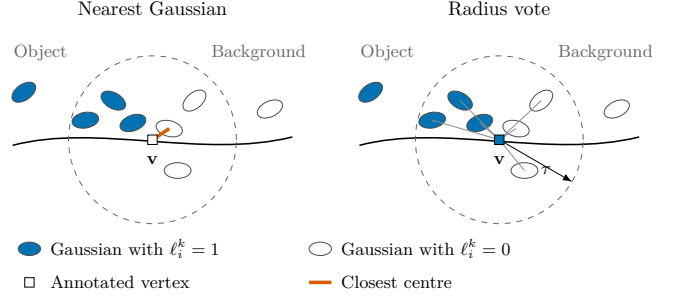

The second operator, which we call radius vote, is a weighted vote inside the neighbourhood. In this case, each Gaussian of the neighbourhood votes with a weight
that decreases with the squared distance and increases with the opacity,
\begin{equation}
  \nu_i(\mathbf{v})=\frac{\max(o_i,o_{\min})}{\lVert\boldsymbol{\mu}_i-\mathbf{v}\rVert^{2}+\epsilon},
  \label{eq:vertexweight}
\end{equation}
where $o_{\min}$ prevents an almost transparent Gaussian from losing its vote
completely, and $\epsilon=10^{-10}$ keeps the quotient finite when a centre
falls on the vertex itself. The votes are split between the Gaussians that carry the
class and the ones that do not,
\begin{equation}
  \begin{split}
    S^{+}(\mathbf{v})&=\sum_{i\in\mathcal{N}(\mathbf{v})}\ell_i^{k}\,\nu_i(\mathbf{v}),\\
    S^{-}(\mathbf{v})&=\sum_{i\in\mathcal{N}(\mathbf{v})}\bigl(1-\ell_i^{k}\bigr)\,\nu_i(\mathbf{v}),
  \end{split}
  \label{eq:vertexmass}
\end{equation}
and the vertex takes the label of class $k$ when the fraction of the first sum over both of them
reaches a threshold $\theta$, as can be seen in Figure~\ref{fig:vertexvote}:
\begin{equation}
  \frac{S^{+}(\mathbf{v})}{S^{+}(\mathbf{v})+S^{-}(\mathbf{v})}\ge\theta.
  \label{eq:vertexrule}
\end{equation}

The term $S^{-}$ is a second form of background competition, apart from the $E_i^{-}$ channel
of Equation~\eqref{eq:evidence}. Without it, every vertex within $\tau$ of a selected Gaussian
would take the class, since in a one-vs-rest run there are no Gaussians of other classes to compete with.
Note that a value of $\theta$ below $0.5$ lets the class win a vertex where the background holds
more weight, so the labelled region can grow beyond the selected Gaussians, up to a distance $\tau$.

The two operators share $\tau$, while $\theta$, $o_{\min}$ and the opacity weighting only exist
in the radius vote. These parameters were fixed on the development scenes, and the operator itself
is chosen on the validation scenes, together with $(\beta,\gamma)$, as Section~\ref{sec:selection}
explains. The rule finally chose the nearest Gaussian.

As for the ground-truth transfer reference, it follows the same path, but starting from the mesh annotation. First,
the annotation is transferred to the Gaussians with the radius vote, where every vertex within
$\tau$ votes with uniform weight, since a vertex has no opacity, and every label of the dataset
competes. Then, these Gaussian labels go back to the mesh with the same operator as the prediction.
The reference and the prediction only differ in where the Gaussian labels come from, so the
reference measures what the representation and the transfer lose, and it depends on the
Gaussian model, on the mesh and on the transfer settings.

%% file: sections/implementation.tex
\section{Implementation and reproducibility}
\label{sec:implementation}

The pipeline has been implemented taking reproducibility into account. The stages run inside container
images with fixed versions, one for training, one for lifting and one for COLMAP \cite{schoenberger2016sfm}. In this way, the versions of CUDA, PyTorch \cite{paszke2019pytorch} and
COLMAP are the same in all the experiments reported here. The host launches these containers
and takes care of loading the scenes, the metrics, the cache and the analytics. In general terms, the pipeline is a sequence of stages, from the preparation of the dataset to the metrics, and each stage
computes its outputs only from its inputs. The $\beta$ and
$\gamma$ sweeps read the projected votes from the cache instead of computing
them again, so the projection of Section~\ref{sec:method-raster} runs once per scene
and class, independently of how many candidates the sweep has.

The elapsed time and the peak CUDA memory of every run and
of each of its stages were recorded. For each stage we also store whether it computed its outputs
or read them from the cache, so both cases, cache miss and cache hit, are measured
on the same scene and the same hardware. These values are the ones reported in Section~\ref{sec:results-cost}.

The analytics of all the runs are stored in several normalised CSV tables, indexed by run, scene,
class and $\beta$. They contain the parameters used to launch each run, the vertex counts per class of
Section~\ref{sec:setup-support}, the quantiles of the target evidence fraction,
and the counts and metrics behind the numbers of this paper. Moreover, each run saves some execution
metadata together with the results, such as the code commit, the GPU model, the driver version and the
command-line arguments.

%% file: sections/evaluation_protocol.tex
\section{Experimental setup and evaluation protocol}
\label{sec:setup}

\subsection{Datasets and scene splits}
\label{sec:setup-data}

We evaluate the method on two public datasets of indoor scenes, Replica and ScanNet++, which are
widely used in this area. Replica contains synthetic indoor scenes. We use its semantic meshes
\cite{replica19arxiv,replicarepo} and the pre-rendered sequences released
with Semantic-NeRF \cite{zhi2021semantic,semanticnerfrepo}. The
cameras are pinhole, with a resolution of \(640\times480\), focal lengths
\(f_x=f_y=320\) and the principal point at the image centre. In order to reduce the computation time,
we subsample each sequence by a factor of five, so only one of every five frames is used in
the lifting. The poses are the trajectories that come with the sequences,
but we convert them to the COLMAP text format and add a point cloud sampled from
the depth frames, so the Gaussian model is trained with those extrinsics.

ScanNet++, on the other hand, contains real indoor scans with geometry from a laser scanner and a denser semantic
annotation \cite{yeshwanth2023scannetpp,scannetpprepo}. Its DSLR images are undistorted
with COLMAP \cite{schoenberger2016sfm} using the reconstruction provided by the dataset,
with at most \(1600\) pixels on the long side, and every camera of that model is used.
The principal point of Equation~\eqref{eq:projection} is the one given by
COLMAP. With respect to the training resolution, for Replica we train the Gaussian model at the full frame size, while for ScanNet++
we train it at half the undistorted resolution.

The ground truth is not annotated in the same way in both datasets, so the preprocessing is also slightly different.
Replica labels the faces of its mesh, whereas ScanNet++ labels the
vertices directly. In Replica, a vertex takes the label of the majority of the faces that meet at it,
as long as that majority reaches \(0.6\) of them.
The vertices below this fraction are left unannotated and are never evaluated. Also, Replica renders its
sequences from the labelled mesh, so the geometry, the annotation and the images are the same object,
while in ScanNet++ they are three independent measurements of the room.

The pre-rendered version of Replica that we use \cite{zhi2021semantic} has eight scenes. We used
\texttt{office\_0} as the development scene during the implementation, and the other seven
scenes form the validation split, where the sweeps are carried out and from which the
scene means and the main tables are obtained. From ScanNet++ we take a held-out test split:
\texttt{7831862f02} was used for development, and ten other
scenes form the test set. These are the downloaded scenes with the most evaluated classes
in their 3D annotation, breaking ties by the scene identifier. This rule only reads the annotation,
and the list was fixed before evaluating any of them. The configuration selected on Replica is applied to them
without retuning it per scene or per class. It should be noted that the development scenes are not included in any validation or test summary.

The names of the classes of each dataset were mapped by hand to the names used by the detector.
Replica has six classes and ScanNet++ has seven, and four of them are
common to both. Bench only appears in the ScanNet++ development scene, so the
test scenes are evaluated on the other six. In the case of ScanNet++, whose vocabulary is much larger, one class of the detector is often mapped to several classes
of the dataset. In each scene, the evaluated classes are those of
the mapped vocabulary that appear in at least one 2D mask derived from the dataset, and this
set is the same for both mask sources. If a class is present in a scene
but no sampled camera gives a mask of it, the class is not evaluated in that scene.
A class that the detector never proposes is treated as an empty prediction,
so its misses are counted and not removed from the scene.

\subsection{Two-dimensional mask sources}
\label{sec:setup-masks}

For the detector condition, the \texttt{yolo26x-seg} segmentation checkpoint
from the YOLO26 family \cite{yolo26} is used, through the Ultralytics package \cite{ultralytics}.
Detections with a confidence below \(0.75\) are discarded, and the soft masks of the
remaining ones are binarised at \(0.5\). This confidence threshold was chosen after a small sweep:
at \(0.5\) too many weak detections entered the vote, and at \(0.75\) most of them
were removed.

In the annotation-derived condition, these masks are replaced by masks rendered from the
dataset labels, with the same format and confidence one on every annotated
pixel. Replica already provides them, since the Semantic-NeRF sequences include a semantic frame
next to each colour frame, so in this dataset the condition works as a 2D ground truth. ScanNet++, however,
does not have these frames, and its masks are obtained by
rasterising the annotated 3D mesh into each camera with nvdiffrast
\cite{laine2020nvdiffrast}, following the official ScanNet++ toolkit \cite{scannetpprepo}.
Running the same lifting under both conditions allows us to separate the errors of the detector from those
of the lifting, and this is what the detector gap of Section~\ref{sec:setup-gaps} measures.

\subsection{Evaluation support}
\label{sec:setup-support}

The predictions are evaluated on the annotated vertices of the scene mesh that the
sampled cameras have seen. With this criterion, vertices behind a wall or in a hidden part of
the room are not counted as misses of the method. The way in which this visibility is decided is different in each dataset, as it depends on the data that each one provides. A Replica vertex
is visible when it is in front of a camera, it projects inside the image and its
depth agrees, within $0.05$ metres, with the depth frame of the sequence at that pixel.
For ScanNet++, the annotated mesh is rasterised with nvdiffrast
\cite{laine2020nvdiffrast}, and the three vertices of a triangle are marked as
visible when the triangle covers at least one pixel.

\subsection{Metrics and aggregation}
\label{sec:setup-metrics}

For each class and scene we report the one-vs-rest IoU, as well as the precision
and the recall, which are the usual metrics in semantic segmentation. A vertex counts as annotated when its label belongs to the dataset vocabulary,
whether or not that class is evaluated. This means that the negatives of a class are annotated surfaces and
not unlabelled space. A class with annotated positives but no predicted
positive gets an IoU of zero.

Apart from these metrics, the predictions are also compared with the ground-truth transfer reference.
Let $\text{IoU}_k^{\text{R}}$ be the IoU obtained when the ground-truth
annotations are transferred through the Gaussian model and back to the mesh.
When this reference is zero, not even the annotation can reach the class through the model. For this reason, every summary only uses the pairs with a positive reference. Each scene gives one value per class,
the mIoU of a scene is the mean over those classes, and the result of a split is the mean over its scenes, together with the standard deviation between
scenes. Since a split has only seven or ten scenes, we also give a 95\% confidence interval for that mean
with a percentile bootstrap over scenes: the scenes are resampled with replacement ten thousand times, and the interval
holds the central 95\% of the resampled means. When two configurations are compared, the same bootstrap
is applied to their difference scene by scene. For the same pairs we compute
\begin{equation}
  \mathrm{IoU}^{\mathrm{rel}}_{k}
  =\frac{\mathrm{IoU}_{k}}{\mathrm{IoU}^{\mathrm{R}}_{k}},
  \label{eq:relative}
\end{equation}
and the relative mIoU, $\mIoU_{\mathrm{rel}}$, is its mean. The relative IoU of a class could be greater
than one when the lifting reaches annotated vertices that the transferred annotation does not reach.

\subsection{Error decomposition}
\label{sec:setup-gaps}

One of the main goals of this work is to know how much each part of the pipeline contributes to
the final error. For this reason, we use three quantities to separate the sources of error. Let
\(\mIoU_{\mathrm{D}}\), \(\mIoU_{\mathrm{A}}\) and \(\mIoU_{\mathrm{R}}\)
be the mIoU obtained with the 2D masks of the detector, with the 2D masks from the dataset annotations
and with the ground-truth transfer reference, respectively. We define

\begin{equation}
  \begin{split}
    g_{\mathrm{det}} &= \mIoU_{\mathrm{A}}-\mIoU_{\mathrm{D}},\\
    g_{\mathrm{lift}} &= \mIoU_{\mathrm{R}}-\mIoU_{\mathrm{A}},\\
    g_{\mathrm{rep}} &= 1-\mIoU_{\mathrm{R}}.
  \end{split}
  \label{eq:gaps}
\end{equation}

The three quantities depend on the settings of Section~\ref{sec:method-transfer}, and their sum
is $1-\mIoU_{\mathrm{D}}$, which is the total error. As the mIoU gives the same weight
to all the classes of the dataset, although some classes cover more mesh vertices than others, 
two errors of the same size do not necessarily correspond to the same number of mislabelled vertices.

%% file: sections/experimental_design.tex
\section{Experimental design}
\label{sec:design}

\subsection{Frozen configuration}
\label{sec:design-frozen}

Apart from the operating point, the method has several other parameters, and they were fixed on two
development scenes, \texttt{office\_0} from Replica and \texttt{7831862f02} from
ScanNet++. The most important ones were the transfer radius and the minimum vote fraction,
since together they decide how the filtered Gaussians interact with the mesh.

These two parameters were swept separately, with the radius vote and with the masks derived from the dataset. First, with \(\theta\) fixed at \(0.5\), the default value of the implementation,
we ran \(\tau\) over \(\{0.02, 0.03, 0.05, 0.08, 0.10, 0.15, 0.20\}\) metres. Then, with \(\tau\)
fixed at \tauStar, we ran \(\theta\) over \(\{0.3, 0.4, 0.5, 0.6, 0.7\}\).
Each candidate was scored by the mean \(\mIoU\) of the two development scenes,
and among the candidates whose mean was within \(0.01\) of the best one, we chose the one where the
two scenes differed the least. This gave $\tau=\tauStar$~m and $\theta=\thetaStar$. Both sweeps
are shown in Figure~\ref{fig:development} of the appendix. It should be noted that no validation or
test scene was used during this procedure.

\begin{table}[t]
\centering
\caption{Configuration fixed before the validation sweep. The last four rows only apply to the radius
vote.}
\label{tab:config}
\vspace{6pt}
\begin{tabular}{@{}lr@{}}
\toprule
Parameter & Value \\
\midrule
\multicolumn{2}{@{}l}{\emph{Model}}\\
Gaussian model iteration            & $30000$ \\
\addlinespace[3pt]
\multicolumn{2}{@{}l}{\emph{Masks}}\\
Detector score threshold            & $0.75$ \\
Mask binarisation                   & $0.5$ \\
\addlinespace[3pt]
\multicolumn{2}{@{}l}{\emph{Voting}}\\
Fallback confidence $q_0$           & $0.25$ \\
Contributing views                  & class present \\
Tile size                     & $16$ px \\
\addlinespace[3pt]
\multicolumn{2}{@{}l}{\emph{Hysteresis}}\\
Connectivity radius $\eta$          & $0.05$ m \\
\addlinespace[3pt]
\multicolumn{2}{@{}l}{\emph{Transfer}}\\
Operator                            & chosen on validation \\
Radius $\tau$                       & \tauStar{} m \\
Minimum vote fraction $\theta$      & \thetaStar{} \\
Opacity floor $o_{\min}$            & $0.1$ \\
Opacity weighting                   & enabled \\
Background competition              & both directions \\
\bottomrule
\end{tabular}
\end{table}

From this point on, the parameters of Table~\ref{tab:config} were kept fixed for the rest of the experiments.
The seven Replica validation scenes were used to choose the operating point, and the ten
held-out ScanNet++ scenes were evaluated with everything already fixed, without retuning
anything per scene or per class.

\subsection{Operating point selection}
\label{sec:selection}

A candidate is a transfer operator together with a pair $(\beta,\gamma)$, where the two operators are
the ones of Section~\ref{sec:method-transfer}. The grid has thirteen values of $\beta$,
\begin{equation*}
  \begin{split}
    \beta \in \{\;&0.50,\,0.70,\,0.90,\,0.94,\,0.95,\,0.96,\,0.97,\,0.975,\\
                  &0.98,\,0.985,\,0.99,\,0.995,\,0.999\,\},
  \end{split}
\end{equation*}
denser above $0.9$, where the result changes fastest when there is no hysteresis, and five values
of $\gamma$: $0$, $0.5$, $0.7$, $0.8$ and $0.9$, so in total there are \candidateCount{} candidates.
All of them were run on the seven validation scenes, with both mask sources, and we applied the following
rule to the results with the masks derived from the dataset.

The idea behind the rule is to prefer a stable candidate over a slightly better but less stable one.
Among the candidates whose mean $\mIoU$ over the
validation scenes is within $0.01$ of the best mean, it takes the one with the smallest
standard deviation between scenes:
\begin{equation}
  \begin{split}
  m^{*} &= \max_{c}\;\operatorname{mean}_{s}\bigl(\mIoU_{s,c}\bigr),\\
  \mathcal{E} &= \bigl\{c:\operatorname{mean}_{s}
                  \bigl(\mIoU_{s,c}\bigr)\geq m^{*}-0.01\bigr\},\\
  \hat{c} &= \operatorname*{arg\,min}_{c\in\mathcal{E}}\;
       \operatorname{std}_{s}\bigl(\mIoU_{s,c}\bigr),
  \end{split}
  \label{eq:selection}
\end{equation}
where $c$ goes over the candidates and $s$ over the seven scenes. We break possible ties using the radius vote,
then the smaller $\beta$ and finally the smaller $\gamma$. The rule chose the \transferStar{} with
$(\beta,\gamma)=(\betaStar,\gammaStar)$. We also apply the same rule to the radius vote alone, which gives
$(\radiusBetaStar,\radiusGammaStar)$, and we report this point on the test scenes too, as this
allows us to see the effect of the transfer operator on its own.

\subsection{Baseline}
\label{sec:baseline}

In order to see the effect of the fraction, we compare the method with the version that we used before
it. That version accumulated the
same target evidence $E_i^{+}$ and selected a Gaussian when
\begin{equation}
  E_i^{+} \ge \beta\,\lvert\mathcal{V}^{k}\rvert,
  \label{eq:baseline}
\end{equation}
that is, it thresholded the target evidence per view where the class appears. Note that we keep the
name $\beta$ because it plays the same role, since it sets the threshold, but it is not the same
quantity as in the method. The fraction of Equation~\eqref{eq:score} lies in $[0,1]$, while the evidence
per view is a sum of weighted confidences over many pixels, so this $\beta$ does not have to be between
zero and one. Everything else is the same as in the method: the votes, the hysteresis with the same
values of $\gamma$, the nearest Gaussian for the transfer and the value of $\tau$. The threshold has no
natural scale here, so its grid has thirteen values between $0.003$ and $1$, around the values that
this version used, and the rule of
Equation~\eqref{eq:selection} chooses its pair $(\beta,\gamma)$ on the same validation scenes. In this
way, the comparison only changes the score that the threshold cuts.

\subsection{Sensitivity ablations}
\label{sec:ablations}

The opacity weighting and the background competition of the transfer only exist in the radius
vote, so the ablations change one factor at a time around the radius vote at its own selected
point, $(\radiusBetaStar,\radiusGammaStar)$, on the validation scenes and with the masks derived
from the dataset. In particular, we tested the following factors:

\begin{itemize}[leftmargin=*,itemsep=2pt]
  \item disabling the hysteresis, that is, setting $\gamma=0$.
  \item disabling the background competition in the transfer, in each direction separately and in both at the same time.
  \item disabling the opacity weighting.
  \item using all the matched cameras instead of only the ones that contain the class, as in Equation~\eqref{eq:views}.
\end{itemize}

Every ablation uses the same scenes and classes as its baseline, so we compare them scene by scene.

\subsection{Detector masks against the 3D result}
\label{sec:statistics}

In order to know what the fusion of views adds to the detector, we also measure, for every class and scene, the
pixel IoU, precision and recall of the YOLO masks against the annotation masks over all the views, and
we compare them with the 3D IoU that the method reaches from the same YOLO masks.

%% file: sections/results.tex
\section{Results}
\label{sec:results}

\subsection{Results at the selected point}
\label{sec:results-main}

\begin{table*}[t]
\centering
\caption{Results at the selected point. The mIoU is the mean over scenes with its standard deviation,
and the 95\% confidence interval comes from the bootstrap of Section~\ref{sec:setup-metrics}.}
\label{tab:main-results}
\vspace{6pt}
\begin{tabular}{@{}llrrrrrrr@{}}
\toprule
Dataset & Masks & Scenes & mIoU & 95\% CI & Precision & Recall & Reference & $\mIoU_{\mathrm{rel}}$ \\
\midrule
\input{tables/main_results} \\
\bottomrule
\end{tabular}
\end{table*}

\begin{figure*}[t]
\centering
\figslot{\textwidth}{0.49\textwidth}{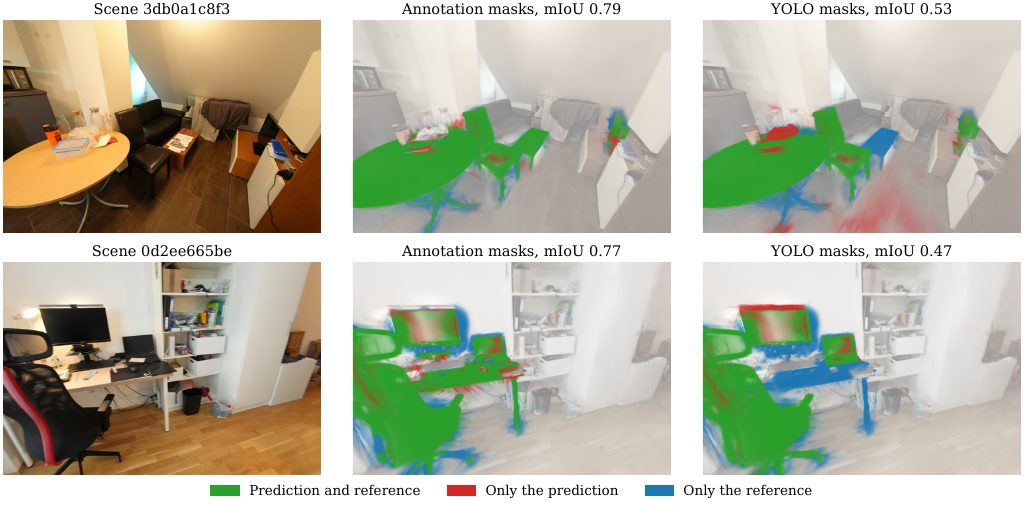}
\caption{Prediction against reference in two ScanNet++ test scenes close to the median, over all their
evaluated classes. The mIoU is the one of the whole scene, measured on the mesh.}
\label{fig:qualitative}
\end{figure*}

Table~\ref{tab:main-results} shows the results at the point selected on validation. In general, the
results follow the same pattern in both datasets, although they are lower on the real scans of ScanNet++. On the ten
held-out ScanNet++ scenes, the mean mIoU is \scannetGTmIoU{} with the annotation masks, with a 95\% confidence interval
from \scannetGTCILow{} to \scannetGTCIHigh{}, and \scannetYOLOmIoU{} with YOLO masks. In
other words, a single operating point, chosen on synthetic rooms, is still valid for the real scans of another dataset without retuning anything. With the annotation masks, precision and recall stay close
to each other, so the prediction does not go beyond the objects, and it does not leave out large parts of them either.

\subsection{Two test scenes}
\label{sec:results-qualitative}

Figure~\ref{fig:qualitative} shows, as an example, two test scenes whose mIoU is close to the median. With the
annotation masks, most of the errors appear as thin bands at the borders of the objects, where the mesh
and the Gaussians do not agree on the position of the surface. With YOLO masks, on the other hand, the errors depend mostly on the detector. The round
table of the first scene looks like a COCO dining table and is well labelled, but the coffee table next
to the sofa and the desk of the second scene are blue, since YOLO does not propose them. The red
Gaussians on the round table are objects that lie on it and enter the mask of the detector.

\subsection{Against the evidence per view}
\label{sec:results-baseline}

\begin{table}[t]
\centering
\caption{mIoU of the baseline and of the method, each one at the point that the same rule chose for it.}
\label{tab:baseline}
\vspace{6pt}
\resizebox{\columnwidth}{!}{%
\begin{tabular}{@{}lrrrrr@{}}
\toprule
& & \multicolumn{2}{c}{Replica} & \multicolumn{2}{c}{ScanNet++} \\
\cmidrule(lr){3-4}\cmidrule(l){5-6}
Thresholded score & $(\beta,\gamma)$ & annotation & YOLO & annotation & YOLO \\
\midrule
\input{tables/baseline} \\
\bottomrule
\end{tabular}}
\end{table}

\begin{figure}[t]
\centering
\figslot{\columnwidth}{0.55\columnwidth}{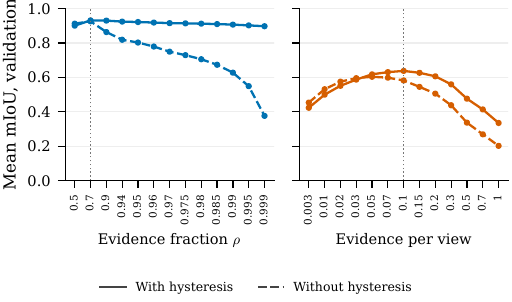}
\caption{Mean validation mIoU along the threshold grid of each score, with the annotation masks. The
dotted line is the selected threshold.}
\label{fig:thresholds}
\end{figure}

Table~\ref{tab:baseline} compares the method with the baseline of Section~\ref{sec:baseline}, and the
fraction is better in every configuration. On the test scenes, the difference is
\baseDeltaScannetGT{} with the annotation masks, with an interval from \baseCILowScannetGT{} to
\baseCIHighScannetGT{}, and the method is better in all \baseScenesScannetGT{} scenes. With YOLO masks
the difference is \baseDeltaScannetYOLO{}, from \baseCILowScannetYOLO{} to \baseCIHighScannetYOLO{}.

Figure~\ref{fig:thresholds} shows the difference between both methods, as well as the stability that the hysteresis brings.
With hysteresis, the fraction gives
almost the same result along its whole grid, while the evidence per view has a narrow peak and goes
from \baseRangeLow{} to \baseRangeHigh{}. The best threshold of the baseline also changes with the class, from
\baseClassBetaMin{} to \baseClassBetaMax{}. The reason why this happens is that the evidence still grows with the size of the object
in the image, with its distance to the camera and with the occlusions. In addition, a background Gaussian that many
views see collects target evidence at the borders of the masks, so the baseline selects too much: on
the test scenes its precision is \baseScannetGTPrec{} for a recall of \baseScannetGTRec{}. The fraction,
instead, compares the target evidence with all the evidence that reached the Gaussian, so these factors cancel.

\subsection{Where the error comes from}
\label{sec:results-gaps}

\begin{table}[t]
\centering
\caption{The three terms of Equation~\eqref{eq:gaps} at the selected point.}
\label{tab:gaps}
\vspace{6pt}
\begin{tabular}{@{}lrrr@{}}
\toprule
Dataset & $g_{\mathrm{det}}$ & $g_{\mathrm{lift}}$ & $g_{\mathrm{rep}}$ \\
\midrule
Replica   & \replicaDetectorGap{} & \replicaLiftingGap{} & \replicaRepGap{} \\
ScanNet++ & \scannetDetectorGap{} & \scannetLiftingGap{} & \scannetRepGap{} \\
\bottomrule
\end{tabular}
\end{table}

\begin{figure}[t]
\centering
\figslot{\columnwidth}{0.95\columnwidth}{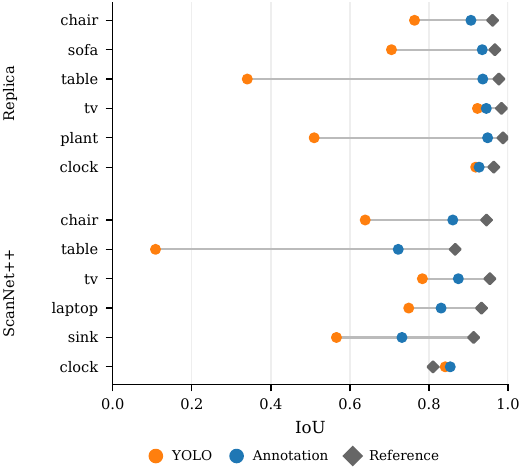}
\caption{IoU of each class with each mask source and for the reference, averaged over the scenes that
contain the class.}
\label{fig:class-gaps}
\end{figure}

\begin{figure}[t]
\centering
\figslot{\columnwidth}{0.9\columnwidth}{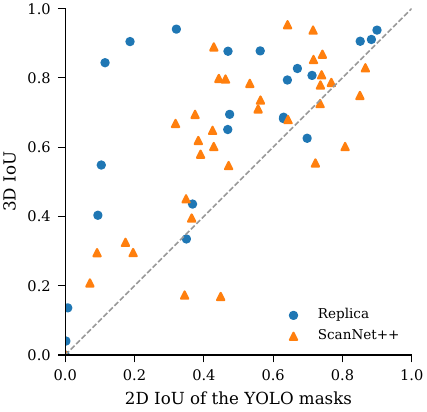}
\caption{Pixel IoU of the YOLO masks against the 3D IoU that the method reaches from them, one marker
per class and scene. Above the diagonal, the 3D result is better than the masks.}
\label{fig:mask-2d-3d}
\end{figure}

Table~\ref{tab:gaps} splits the error following Section~\ref{sec:setup-gaps}. The detector is the
largest source of error in both datasets, with \scannetDetectorGap{} on ScanNet++, while the lifting
loses \scannetLiftingGap{}. Therefore, most of what can still be gained is in the 2D masks, and with the
annotation masks the method already reaches \scannetGTrel{} of the reference on ScanNet++.

Figure~\ref{fig:class-gaps} shows that the detector gap is not spread evenly over the classes. The clearest
case is the one of the table class. On ScanNet++, its IoU is \clsScannetTableGT{} with the annotation masks and
\clsScannetTableYOLO{} with YOLO, which gives it zero in \clsScannetTableZero{} of the
\clsScannetTableScenes{} scenes. The detector was trained on COCO, whose only table is the dining
table, while most tables of ScanNet++ are desks, so YOLO recovers only \maskScannetTableTwoDRec{} of the
table pixels and there is almost nothing to lift. As could be expected, the representation term is larger on ScanNet++, and
the clock is the worst case, with a reference of \clsScannetClockRef{}, since it is a small object
covered by few Gaussians in real scans.

\subsection{What the fusion of views adds to the detector}
\label{sec:results-fusion}

Figure~\ref{fig:mask-2d-3d} compares, for every class and scene, the YOLO masks in 2D with the 3D
result that the method obtains from them. The 3D IoU is higher in \replicaMaskWins{} of the
\replicaMaskPairs{} pairs of Replica and in \scannetMaskWins{} of the \scannetMaskPairs{} pairs of
ScanNet++, and on ScanNet++ the mean goes from \scannetMaskTwoD{} to \scannetMaskThreeD{}. In 2D, YOLO is relatively precise but it misses a large part of the objects, with a pixel precision of \scannetMaskTwoDPrec{} and a recall of
\scannetMaskTwoDRec{} on ScanNet++. When it misses an object in some views, the Gaussians of the object
still collect target evidence from the views where it finds it, and a view without the class does not
contribute at all. An example of this can be seen in Figure~\ref{fig:overview}, with the tables of a classroom. Nevertheless, the desks,
which YOLO almost never proposes, stay low in 3D as well.

\subsection{The operating point}
\label{sec:results-stability}

\begin{figure}[t]
\centering
\figslot{\columnwidth}{1.0\columnwidth}{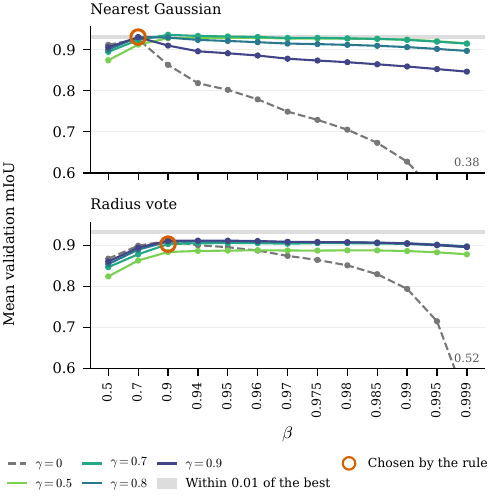}
\caption{Mean validation mIoU along $\beta$ for every $\gamma$, with the annotation masks. The grey band
holds the candidates within $0.01$ of the best mean.}
\label{fig:beta-gamma}
\end{figure}

Figure~\ref{fig:beta-gamma} shows the results of the whole validation grid. Of the \candidateCount{} candidates,
\eligibleCount{} were within $0.01$ of the best mean, \bestMean{}, and all of them use the nearest
Gaussian. Among them, the rule chose $(\betaStar,\gammaStar)$, with a mean of \selectedMean{} and a
standard deviation between scenes of \selectedSd{}.

The figure also allows us to see the effect of the hysteresis. Without it, in the dashed curves, the mean
goes down from \noHystRangeHigh{} to \noHystRangeLow{} as $\beta$ grows. With $\gamma=\gammaStar$,
it stays between \hystRangeLow{} and \hystRangeHigh{} for every $\beta$ of the grid. So, the
hysteresis does not raise the best result, but it makes the result almost independent of $\beta$: a
high $\beta$ keeps only the clearest Gaussians as seeds, and the hysteresis recovers the rest of the
object through the Gaussians connected to them. The same effect is shown class by class in
Figure~\ref{fig:beta-curves} of the appendix.

\subsection{The transfer operator}
\label{sec:results-operator}

\begin{table*}[t]
\centering
\caption{Each transfer operator at the point that the rule chooses for it.}
\label{tab:operators}
\vspace{6pt}
\begin{tabular}{@{}llrrrrrr@{}}
\toprule
& & \multicolumn{3}{c}{Replica, validation} & \multicolumn{3}{c}{ScanNet++, test} \\
\cmidrule(lr){3-5}\cmidrule(l){6-8}
Operator & $(\beta,\gamma)$ & annotation & YOLO & reference & annotation & YOLO & reference \\
\midrule
\input{tables/operators} \\
\bottomrule
\end{tabular}
\end{table*}

Table~\ref{tab:operators} compares the two operators, each one at its own selected point. On
validation, the nearest Gaussian is better by \opGainReplicaGT{} with the annotation masks, and its
dispersion between scenes is a third of the one of the radius vote. On the test scenes, with the
annotation masks, it is better by \opGainScannetGT{}. With YOLO masks, however, the radius vote is slightly better: the
difference is \opDeltaScannetYOLO{}, with an interval from \opCILowScannetYOLO{} to
\opCIHighScannetYOLO{}.

This is due to the minimum vote fraction of the radius vote. With $\theta=\thetaStar$, the class wins a
vertex with less than half of the vote, so the labelled region grows up to $\tau$ around the selected
Gaussians. On Replica, the clock is a good example: with the radius vote, its precision with the
annotation masks is \radiusClsReplicaClockGTPrec{} for a recall of \radiusClsReplicaClockGTRec{}. Taking
into account that the YOLO masks leave part of the objects out, this growth recovers some recall, and this explains
the small advantage of the radius vote in that condition. The nearest Gaussian does not grow anything,
so the labelled surface simply follows the lifting.

\subsection{Sensitivity ablations}
\label{sec:results-ablations}

\begin{table}[t]
\centering
\caption{Ablations on the validation scenes with the annotation masks. Every row changes one factor of
the radius vote at its selected point, and the difference is paired scene by scene.}
\label{tab:ablations}
\vspace{6pt}
\resizebox{\columnwidth}{!}{%
\begin{tabular}{@{}lrrr@{}}
\toprule
Configuration & mIoU & Difference & Reference \\
\midrule
\input{tables/ablations} \\
\bottomrule
\end{tabular}}
\end{table}

Table~\ref{tab:ablations} shows the results of the ablations. The background competition from Gaussians to mesh is
the only factor that has an important effect on the result. Without it, every vertex within $\tau$ of a selected Gaussian takes
the class, and the mIoU changes by \ablNoGtoMDelta{}. Without competition from mesh to Gaussians, the prediction does not change and only the
reference goes down, to \ablNoMtoGRef{}, since then the Gaussians of other surfaces keep the label of
any class vertex within $\tau$.

The rest of the factors hardly change the result. Without opacity weighting and with every matched
camera, the difference stays below $0.01$. We expected the cameras without the class to
dilute the fraction, but a camera that sees an object also has it in its mask, so these extra cameras
add non-target evidence mostly to Gaussians that do not belong to the class, which were not going to be selected in any case. Finally, disabling the
hysteresis at $\beta=\radiusBetaStar$ gives \ablNoHystDelta{}, which agrees with
Section~\ref{sec:results-stability}: at a moderate $\beta$ the hysteresis is not needed, and the reason why it is useful is that the result does not depend on this choice.

\subsection{Cost}
\label{sec:results-cost}

\begin{table}[t]
\centering
\caption{Cost of each stage for a whole test scene over the \costScenes{} scenes that ran every stage
from scratch. The time is the median and the memory the highest peak. The last row runs the thirteen
values of $\beta$ from the cached votes.}
\label{tab:runtime}
\vspace{6pt}
\begin{tabular}{@{}lrr@{}}
\toprule
Stage & Time & Peak memory \\
\midrule
Dataset preparation       & \timePrepare{}   & -- \\
Mask generation           & \timeMasks{}     & \memMasks{} \\
Model training            & \timeTraining{}  & \memTraining{} \\
Vote accumulation         & \timeVotes{}     & \memVotes{} \\
Threshold and hysteresis  & \timeThreshold{} & -- \\
Mesh transfer and metrics & \timeTransfer{}  & -- \\
\addlinespace[3pt]
Sweep over $\beta$, cached & \timeSweepWarm{} & -- \\
\bottomrule
\end{tabular}
\end{table}

The last aspect that we analyse is the computational cost of the method. Table~\ref{tab:runtime} shows the cost of a test scene. The median total time of a scene is \timeMissTotal{}, and the
vote accumulation is the slowest stage, with \timeVotes{}, more than the training of the model itself. This is due to the fact that the visibility is computed in PyTorch. The cache is what makes the sweeps
possible: once the votes are stored, the thirteen values of $\beta$ take \timeSweepWarm{}, and no
candidate projects the Gaussians again. All the experiments ran on \hardwareDescription{} GPUs of
the Picasso supercomputer. The results of every class and every scene can be found in Appendix~\ref{app:per-class}.

%% file: tables/main_results.tex
Replica & annotation & 7 & 0.93 {\footnotesize$\pm$0.01} & [0.92, 0.94] & 0.96 & 0.97 & 0.97 & 0.956 \\
 & YOLO & 7 & 0.65 {\footnotesize$\pm$0.13} & [0.55, 0.74] & 0.72 & 0.79 & 0.97 & 0.668 \\
\addlinespace[3pt]
ScanNet++ & annotation & 10 & 0.80 {\footnotesize$\pm$0.05} & [0.77, 0.83] & 0.86 & 0.92 & 0.91 & 0.876 \\
 & YOLO & 10 & 0.54 {\footnotesize$\pm$0.09} & [0.48, 0.60] & 0.64 & 0.71 & 0.91 & 0.582

%% file: tables/baseline.tex
Target evidence per view & $(0.1,\,0.5)$ & 0.64 {\footnotesize$\pm$0.04} & 0.44 {\footnotesize$\pm$0.07} & 0.56 {\footnotesize$\pm$0.05} & 0.42 {\footnotesize$\pm$0.04} \\
Target evidence fraction & $(0.7,\,0.8)$ & 0.93 {\footnotesize$\pm$0.01} & 0.65 {\footnotesize$\pm$0.13} & 0.80 {\footnotesize$\pm$0.05} & 0.54 {\footnotesize$\pm$0.09}

%% file: tables/operators.tex
Nearest Gaussian & $(0.7,\,0.8)$ & 0.93 {\footnotesize$\pm$0.01} & 0.65 {\footnotesize$\pm$0.13} & 0.97 & 0.80 {\footnotesize$\pm$0.05} & 0.54 {\footnotesize$\pm$0.09} & 0.91 \\
Radius vote & $(0.9,\,0.7)$ & 0.90 {\footnotesize$\pm$0.04} & 0.63 {\footnotesize$\pm$0.12} & 0.91 & 0.79 {\footnotesize$\pm$0.05} & 0.55 {\footnotesize$\pm$0.07} & 0.84

%% file: tables/ablations.tex
Radius vote at its selected point & 0.90 {\footnotesize$\pm$0.04} & -- & 0.91 \\
\addlinespace[3pt]
Hysteresis disabled & 0.91 {\footnotesize$\pm$0.04} & +0.008 & 0.91 \\
No competition, Gaussian to mesh & 0.75 {\footnotesize$\pm$0.05} & -0.153 & 0.76 \\
No competition, mesh to Gaussian & 0.90 {\footnotesize$\pm$0.04} & +0.000 & 0.76 \\
No competition, both directions & 0.75 {\footnotesize$\pm$0.05} & -0.153 & 0.63 \\
Opacity weighting disabled & 0.90 {\footnotesize$\pm$0.03} & +0.002 & 0.91 \\
Every matched camera contributes & 0.90 {\footnotesize$\pm$0.04} & +0.001 & 0.91

%% file: sections/limitations.tex
\section{Limitations}
\label{sec:limitations}

As with any method, ours has some limitations, and two of them concern the method itself. First, if a class
never appears in any mask,
it receives no evidence and the method cannot recover it. In the same way, if a
surface is covered by very few Gaussians, the reconstruction noise can make the non-target
evidence dominate, and the method will not recover the target class there.

Second, the method cannot label several target classes at the same time.
The same Gaussian can be selected for two classes in two separate runs, but we do not provide
any procedure for the case where each Gaussian needs a single exclusive label. The cost of a scene grows
with the number of classes, and the vocabulary is limited by what the detector proposes and
by what the dataset annotates.

The correspondence between the dataset names and the detector names was written by hand, since each dataset
has only six or seven classes. In a future version, learned embeddings could propose or check
these correspondences, but they would add a source of variation that is not part of this study.

The ground-truth transfer reference is obtained with the operator of Section~\ref{sec:method-transfer}
and changes with it, as Section~\ref{sec:results-operator} shows for the two operators we tried.
Therefore, the reference measures the cost of the representation and the transfer for one specific
operator, and not for the best possible transfer.

Replica and ScanNet++ differ in many aspects apart from their scenes. Their vertices are annotated
in different ways, and the visibility criteria and the origin of the camera poses are also different.
The class sets are not the same either, with six and seven classes and only four of them in common.
We subsample Replica by a factor of five, while in ScanNet++ we use every camera, and the Gaussian
models of each dataset are trained at different image scales. In Replica the images are renders of the
annotated mesh, so the masks, the Gaussians and the evaluated vertices share the same reference frame.
In ScanNet++ the derived masks also agree with the mesh, but they are related to the photographs
through an estimated alignment. The error of this alignment is present in every condition we measure
there, and this is why we choose the operating point on Replica. The change from validation to test
includes all of these differences at once, and our design only isolates the gap due to the mask source.

In Replica, a vertex is left unannotated when its faces disagree beyond the fraction
of Section~\ref{sec:setup-data}. Because of this, part of the boundary between objects
is not evaluated in that dataset.

Our visibility module is written in plain PyTorch instead of custom CUDA kernels,
which makes the method slower than the original rasteriser. In addition, a tile with many Gaussians allocates
an intermediate tensor of size $N_{\text{Gaussians}} \times N_{\text{pixels}}$, which can cause out-of-memory errors.

Finally, we only compare the method with its own conditions and its own reference,
and we did not run any other published method on these scenes. Therefore, the numbers of
Section~\ref{sec:results} do not show how the method compares with published baselines.

%% file: sections/conclusion.tex
\section{Conclusion}
\label{sec:conclusion}

In this work, we have presented a method that labels a Gaussian scene that is already trained,
one target class at a time. Target and non-target evidence are accumulated at the same time,
and each Gaussian receives a fraction of target evidence. A fixed
threshold $\beta$ selects the seeds, and the spatial hysteresis then adds every
Gaussian above $\gamma\beta$ that belongs to a connected component with at least one seed.
For the evaluation, the labels go to the mesh through the nearest Gaussian.

The transfer parameters were fixed on the development scenes, the seven Replica
validation scenes were used to choose the transfer operator and a single pair $(\beta,\gamma)=(\betaStar,\gammaStar)$,
and the result was applied to the ten ScanNet++ test scenes without retuning anything.
On the validation split, the mean mIoU was \replicaGTmIoU{} with the annotation masks and
\replicaYOLOmIoU{} with YOLO masks, while on the held-out scenes it was
\scannetGTmIoU{} and \scannetYOLOmIoU{}.

From the analysis of the error, we can draw three main conclusions. First, the detector is the main source of error,
with a gap of \scannetDetectorGap{} on the test scenes, and most of it comes from classes that its
vocabulary does not cover well, such as the desks of ScanNet++. Second, the evidence fraction is what
makes a single threshold possible. Compared with the evidence per view of a previous version, it improves the
test mIoU by \baseGainScannetGT{} with the annotation masks and by \baseGainScannetYOLO{} with YOLO
masks, and its threshold moves from Replica to ScanNet++ almost without loss. Third, the hysteresis does not raise the best result, but it makes the
result almost independent of $\beta$, which matters when the same value has to work in another dataset.

Regarding future work, several lines follow from the limitations above. In the first place, a comparison with other published methods on these scenes would allow us to put our results in context. In addition, the results suggest improving the 2D masks before anything else, for example with an open-vocabulary
detector, as well as measuring the coverage of the representation before transferring labels.

%% file: sections/contributions.tex
\newpage
\section*{Author Contributions}
\label{sec:contributions}

The contribution of each author follows the CRediT roles, where I.V.G. is Iván Verdugo Guerra, E.L.R. is Ezequiel López Rubio and J.G.G. is Jorge García González.
\begin{itemize}[leftmargin=*,noitemsep,topsep=2pt,itemsep=1.5pt]
  \item \textbf{Conceptualization:} I.V.G., E.L.R., J.G.G.
  \item \textbf{Methodology:} I.V.G., E.L.R., J.G.G.
  \item \textbf{Software:} I.V.G.
  \item \textbf{Validation:} I.V.G.
  \item \textbf{Formal analysis:} I.V.G.
  \item \textbf{Investigation:} I.V.G.
  \item \textbf{Data curation:} I.V.G.
  \item \textbf{Writing -- original draft:} I.V.G.
  \item \textbf{Writing -- review \& editing:} I.V.G., E.L.R., J.G.G.
  \item \textbf{Visualization:} I.V.G.
  \item \textbf{Supervision:} E.L.R., J.G.G.
  \item \textbf{Project administration:} E.L.R., J.G.G.
  \item \textbf{Resources:} E.L.R., J.G.G.
  \item \textbf{Funding acquisition:} E.L.R., J.G.G.
\end{itemize}

\section*{Acknowledgements}
\label{sec:acknowledgements}

The experiments of this work were run on the Picasso supercomputer. For this reason, the authors thankfully acknowledge the computer resources, technical expertise and assistance provided by the SCBI, the Supercomputing and Bioinformatics center of the University of Malaga.

\section*{Declaration on the use of AI tools}
\label{sec:ai-declaration}

The main parts of the work were carried out by the authors, namely the method formulation and implementation, 
including the evidence fraction and thresholding, 
designing the evaluation protocol and the error decomposition, as well as the execution of the experiments and 
the analysis of the results.

Large language models were used in some specific tasks. First, Claude Opus 5 helped to decide how everything would
be organised in different sections. The writing of the manuscript was then carried out by the first author, and 
afterwards, Claude Opus 5.5 was used to help fix English inconsistencies, while the work was revised by all the 
authors, who take full responsibility for the content of this paper. Finally, DeepSeek V4.1 Flash helped to write 
the code of the folder \texttt{evaluation/scripts}, which generates the figures and tables of the preprint, and Gemini 3.1 Pro 
helped with the code that adapts Replica and ScanNet++ to the pipeline, in the folders \texttt{evaluation/replica} 
and \texttt{evaluation/scannetpp}.

We are in favour of a responsible use of AI tools in research, in which their use is declared openly
and the authors remain responsible for every result.

%% file: sections/appendix.tex
\section{Results by class and by scene}
\label{app:per-class}

Table~\ref{tab:per-class} gives the IoU of each class at the selected point, and
Table~\ref{tab:per-scene} gives the mIoU of every scene with both operators. A class only contributes to
the mean in the scenes that contain it, so the last column of Table~\ref{tab:per-class} shows how much
weight each row has.

\begin{table}[h]
\centering
\caption{IoU per class at the selected point, averaged over the scenes that contain the class.}
\label{tab:per-class}
\vspace{6pt}
\resizebox{\columnwidth}{!}{%
\begin{tabular}{@{}llrrrr@{}}
\toprule
Dataset & Class & Annotation & YOLO & Reference & Scenes \\
\midrule
\input{tables/per_class} \\
\bottomrule
\end{tabular}}
\end{table}

\begin{table*}[t]
\centering
\caption{mIoU of every scene for each operator at its own selected point.}
\label{tab:per-scene}
\vspace{6pt}
\begin{tabular}{@{}llrrrrrr@{}}
\toprule
& & \multicolumn{3}{c}{\TransferStar{}, $(\betaStar,\gammaStar)$} & \multicolumn{3}{c}{Radius vote, $(\radiusBetaStar,\radiusGammaStar)$} \\
\cmidrule(lr){3-5}\cmidrule(l){6-8}
Dataset & Scene & annotation & YOLO & reference & annotation & YOLO & reference \\
\midrule
\input{tables/per_scene} \\
\bottomrule
\end{tabular}
\end{table*}

\section{Development sweep}
\label{app:development}

\begin{figure}[t]
\centering
\figslot{\columnwidth}{0.7\columnwidth}{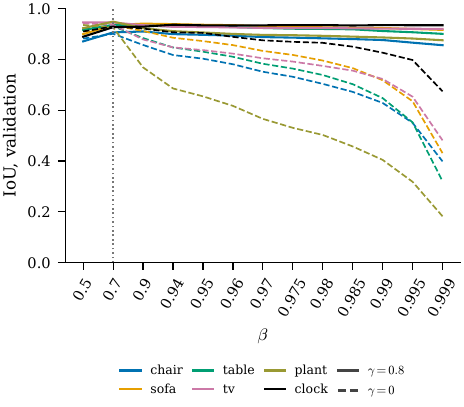}
\caption{IoU of each class on the validation scenes along $\beta$, with the annotation masks, with and
without hysteresis. The dotted line is $\beta=\betaStar$.}
\label{fig:beta-curves}
\end{figure}

\begin{figure*}[t]
\centering
\figslot{0.85\textwidth}{0.3\textwidth}{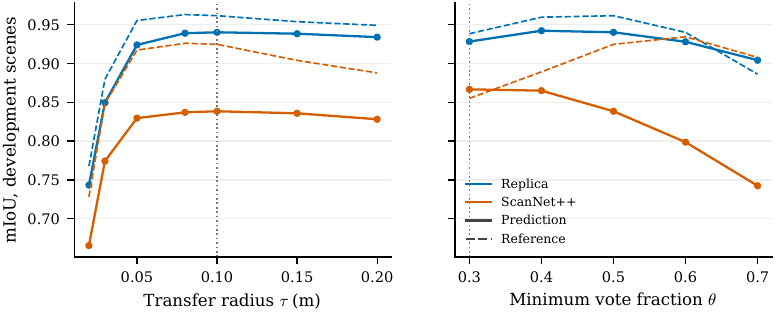}
\caption{Development sweep of the radius vote with the annotation masks. Solid lines are the
prediction and dashed lines the reference.}
\label{fig:development}
\end{figure*}

Figure~\ref{fig:development} shows the results of the two sweeps of Section~\ref{sec:design-frozen}. A small radius
leaves many vertices without any Gaussian, and both the prediction and the reference fall quickly
below $\tau=0.05$~m. Above $0.08$~m the prediction is almost flat. When $\theta$ goes down, the
prediction improves on ScanNet++, while the reference gets worse, since the class takes vertices
where it only has part of the vote. This is the growth that Section~\ref{sec:results-operator}
discusses.

%% file: tables/per_class.tex
Replica & chair & 0.91 & 0.76 & 0.96 & 6 \\
 & sofa & 0.93 & 0.71 & 0.97 & 3 \\
 & table & 0.94 & 0.34 & 0.98 & 6 \\
 & tv & 0.95 & 0.92 & 0.98 & 2 \\
 & plant & 0.95 & 0.51 & 0.99 & 3 \\
 & clock & 0.93 & 0.92 & 0.96 & 3 \\
\addlinespace[3pt]
ScanNet++ & chair & 0.86 & 0.64 & 0.95 & 10 \\
 & table & 0.72 & 0.11 & 0.87 & 9 \\
 & tv & 0.87 & 0.78 & 0.95 & 8 \\
 & laptop & 0.83 & 0.75 & 0.93 & 3 \\
 & sink & 0.73 & 0.57 & 0.91 & 7 \\
 & clock & 0.85 & 0.84 & 0.81 & 2

%% file: tables/per_scene.tex
Replica & \texttt{office\_1} & 0.95 & 0.44 & 0.98 & 0.90 & 0.44 & 0.91 \\
 & \texttt{office\_2} & 0.94 & 0.78 & 0.97 & 0.89 & 0.74 & 0.88 \\
 & \texttt{office\_3} & 0.91 & 0.65 & 0.96 & 0.85 & 0.59 & 0.85 \\
 & \texttt{office\_4} & 0.93 & 0.64 & 0.97 & 0.86 & 0.57 & 0.87 \\
 & \texttt{room\_0} & 0.92 & 0.52 & 0.98 & 0.95 & 0.54 & 0.96 \\
 & \texttt{room\_1} & 0.93 & 0.84 & 0.99 & 0.93 & 0.84 & 0.96 \\
 & \texttt{room\_2} & 0.93 & 0.68 & 0.97 & 0.93 & 0.67 & 0.94 \\
\addlinespace[3pt]
ScanNet++ & \texttt{09c1414f1b} & 0.71 & 0.46 & 0.91 & 0.71 & 0.48 & 0.85 \\
 & \texttt{0d2ee665be} & 0.77 & 0.47 & 0.88 & 0.74 & 0.50 & 0.76 \\
 & \texttt{21d970d8de} & 0.90 & 0.68 & 0.93 & 0.89 & 0.66 & 0.89 \\
 & \texttt{25f3b7a318} & 0.83 & 0.54 & 0.94 & 0.82 & 0.53 & 0.86 \\
 & \texttt{27dd4da69e} & 0.76 & 0.39 & 0.90 & 0.76 & 0.46 & 0.79 \\
 & \texttt{3db0a1c8f3} & 0.79 & 0.53 & 0.91 & 0.76 & 0.57 & 0.85 \\
 & \texttt{3f15a9266d} & 0.81 & 0.63 & 0.88 & 0.78 & 0.59 & 0.80 \\
 & \texttt{5942004064} & 0.82 & 0.67 & 0.93 & 0.81 & 0.69 & 0.87 \\
 & \texttt{5eb31827b7} & 0.87 & 0.55 & 0.94 & 0.86 & 0.56 & 0.88 \\
 & \texttt{6115eddb86} & 0.76 & 0.48 & 0.92 & 0.78 & 0.49 & 0.87